\pdfoutput=1

\documentclass[11pt]{article}

\usepackage[preprint]{acl}

\usepackage{times}
\usepackage{latexsym}

\usepackage[T1]{fontenc}

\usepackage[utf8]{inputenc}

\usepackage{microtype}
\usepackage{inconsolata}
\usepackage{graphicx}
\usepackage{subfigure} 
\usepackage{amsmath}
\usepackage{stfloats}
\usepackage{multirow}
\usepackage{enumitem}
\usepackage{booktabs}
\usepackage{bbding}
\usepackage{amsfonts,amssymb}
\usepackage{diagbox}
\usepackage{newfloat}
\usepackage{listings}
\usepackage{times}  
\usepackage{helvet} 
\usepackage{courier}  
\usepackage{natbib}  
\usepackage{caption}
\usepackage{makecell}
\usepackage{xcolor}
\usepackage{pifont}
\usepackage{soul}
\usepackage{url}
\usepackage{tikz}
\usepackage{colortbl}
\newcommand{\tabincell}[2]{\begin{tabular}{@{}#1@{}}#2\end{tabular}}

\definecolor{best}{RGB}{202,252,209}
\definecolor{second}{RGB}{250, 229, 215}
\definecolor{mention}{RGB}{255, 242, 206}

\title{\texttt{CELLO} \includegraphics[width=0.4cm]{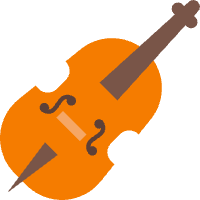}: Causal Evaluation of Large Vision-Language Models}

\author{Meiqi~Chen\footnotemark[1] \\ Peking University \\ \texttt{meiqichen@stu.pku.edu.cn}
        \And
        Bo~Peng \\  Shanghai Jiao Tong University
        \\ Shanghai AI Laboratory
        \\ 
        \texttt{peng\_bo2019@sjtu.edu.cn}
        \\
        \AND
        Yan~Zhang$\footnotemark[2]$ \\ Peking University \\ \texttt{zhyzhy001@pku.edu.cn}
        \And
        Chaochao~Lu$\footnotemark[2]$ \\ Shanghai AI Laboratory \\ \texttt{luchaochao@pjlab.org.cn}
}

\begin{document}
\maketitle
\begin{abstract}
Causal reasoning is fundamental to human intelligence and crucial for effective decision-making in real-world environments.
Despite recent advancements in large vision-language models (LVLMs), their ability to comprehend causality remains unclear.
Previous work typically focuses on commonsense causality between events and/or actions, which is insufficient for applications like embodied agents and lacks the explicitly defined causal graphs required for formal causal reasoning.
To overcome these limitations, we introduce a fine-grained and unified definition of causality involving interactions between humans and/or objects.
Building on the definition, we construct a novel dataset, \texttt{CELLO}, consisting of 14,094 causal questions across all four levels of causality: discovery, association, intervention, and counterfactual. This dataset surpasses traditional commonsense causality by including explicit causal graphs that detail the interactions between humans and objects.
Extensive experiments on \texttt{CELLO} reveal that current LVLMs still struggle with causal reasoning tasks, but they can benefit significantly from our proposed \texttt{CELLO-CoT}, a causally inspired chain-of-thought prompting strategy.
Both quantitative and qualitative analyses from this study provide valuable insights for future research.
Our project page is at {\small \url{https://github.com/OpenCausaLab/CELLO}}.
\end{abstract}

\renewcommand*{\thefootnote}{\fnsymbol{footnote}}
\footnotetext[1]{This work was done during her internship at Shanghai AI Laboratory.}
\footnotetext[2]{Corresponding author.}

\renewcommand*{\thefootnote}{\arabic{footnote}}
\setcounter{footnote}{0} 

\section{Introduction}
\begin{figure}[htbp!]
\centering  
\includegraphics[width=0.45\textwidth]{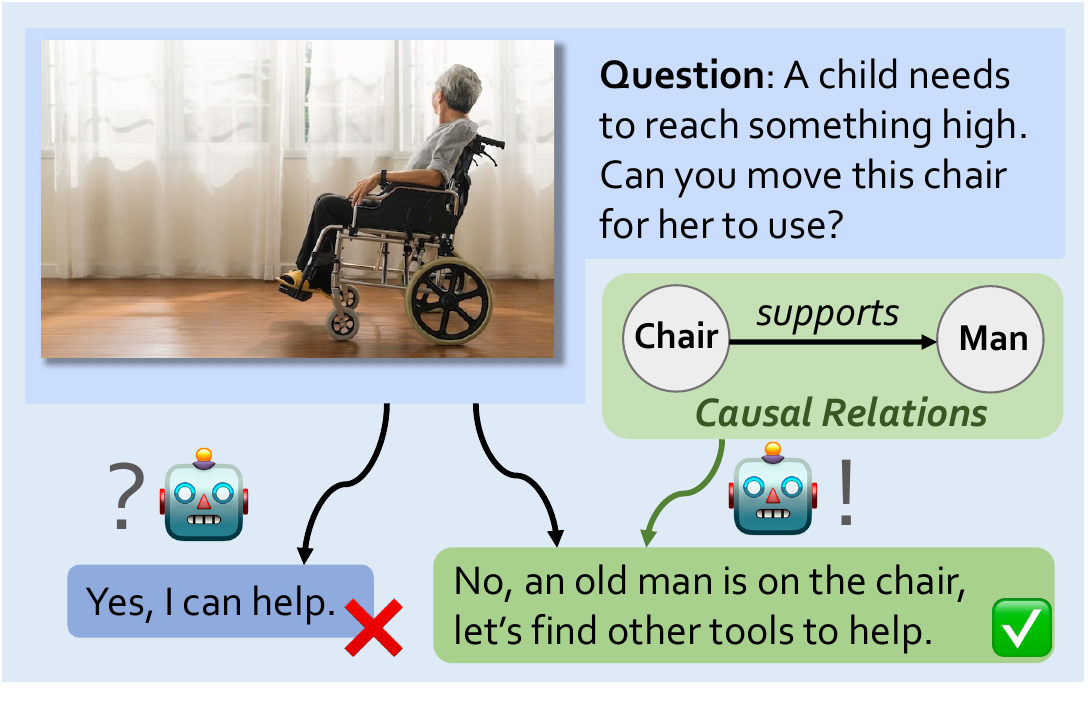}
\vspace{-3mm}
\caption{An example of causal reasoning in the vision-language context. LVLMs (e.g., GPT-4o) might generate inappropriate responses due to a limited understanding of causal relationships.}
\label{fig:example}
\vspace{-5mm}
\end{figure}
Causal reasoning is recognized as a fundamental component of human intelligence~\cite{penn2007causal, harari2014sapiens}. Recent advances in large language models (LLMs) have promoted a surge of research successfully adapting LLMs to vision-language tasks, resulting in powerful large vision-language models (LVLMs)~\cite{gpt2023openai, liu2023visual}. Despite these advancements, a critical question arises: \emph{Do LVLMs really understand causality?}

\begin{table*}
    \renewcommand
    \arraystretch{0.9}
    \centering
    \small
    \setlength{\tabcolsep}{4pt}
    \begin{tabular}{l|cccc|c|c|c|c}
    \toprule
         \multirow{2}{*}{\bf{Datasets}} 
         &\multicolumn{4}{c|}{\bf{Question Types}}
        &\multirow{2}{*}{\bf{\tabincell{c}{Fine-grained \\ Causality}}}
         &\multirow{2}{*}{\bf{Answer Type}}
        & \multirow{2}{*}{\bf{Rationale}}
        &\multirow{2}{*}{\bf{\# Size}}\\ 
        & Disc. & Assoc. &Interv. &CF. & & & \\
 \midrule
         Visual7W ~\cite{zhu2016visual7w}  &\textcolor{yellow}{\ding{51}}&\textcolor{red}{\ding{55}} &\textcolor{red}{\ding{55}} &\textcolor{red}{\ding{55}} &\textcolor{red}{\ding{55}} &Open-ended  &\textcolor{red}{\ding{55}} & 8,884\footnotemark[1]  \\
         VQA (v2)~\cite{goyal2017making}  &\textcolor{yellow}{\ding{51}}&\textcolor{red}{\ding{55}} &\textcolor{red}{\ding{55}} &\textcolor{red}{\ding{55}} &\textcolor{red}{\ding{55}} &Open-ended  &\textcolor{red}{\ding{55}} & 1,952\footnotemark[1]  \\
         FVQA~\cite{wang2017fvqa}  &\textcolor{yellow}{\ding{51}}&\textcolor{red}{\ding{55}} &\textcolor{red}{\ding{55}} &\textcolor{red}{\ding{55}} &\textcolor{red}{\ding{55}} &Open-ended &\textcolor{green}{\ding{52}}  & 17\footnotemark[1] \\
         OKVQA~\cite{marino2019ok}  &\textcolor{yellow}{\ding{51}}&\textcolor{red}{\ding{55}} &\textcolor{red}{\ding{55}} &\textcolor{red}{\ding{55}} &\textcolor{red}{\ding{55}}&Open-ended  &\textcolor{red}{\ding{55}}&115\footnotemark[1]   \\
         VCR~\cite{zellers2019vcr} &\textcolor{yellow}{\ding{51}}&\textcolor{red}{\ding{55}} &\textcolor{red}{\ding{55}} &\textcolor{red}{\ding{55}} &\textcolor{red}{\ding{55}} &Multi-choice  &\textcolor{green}{\ding{52}}&9390\footnotemark[1] \\
         VisualCOMET~\cite{park2020visualcomet} &\textcolor{yellow}{\ding{51}}&\textcolor{red}{\ding{55}} &\textcolor{red}{\ding{55}} &\textcolor{red}{\ding{55}} &\textcolor{red}{\ding{55}} &Open-ended  &\textcolor{green}{\ding{52}}&13,768\footnotemark[1] \\
         BD2BB~\cite{pezzelle-etal-2020-different} &\textcolor{red}{\ding{55}}&\textcolor{red}{\ding{55}} &\textcolor{yellow}{\ding{51}} &\textcolor{red}{\ding{55}}  &\textcolor{red}{\ding{55}} &Multi-choice  &\textcolor{red}{\ding{55}}&10,000 \\
         COSIM~\cite{kim-etal-2022-cosim} &\textcolor{red}{\ding{55}}&\textcolor{red}{\ding{55}} &\textcolor{yellow}{\ding{51}} &\textcolor{red}{\ding{55}}  &\textcolor{red}{\ding{55}} &Multi-choice  &\textcolor{red}{\ding{55}}&3,500 \\
         NORMLENS~\cite{han2023reading} &\textcolor{red}{\ding{55}}&\textcolor{red}{\ding{55}} &\textcolor{yellow}{\ding{51}} &\textcolor{red}{\ding{55}}  &\textcolor{red}{\ding{55}} &Multi-choice  &\textcolor{red}{\ding{55}}&10,000 \\
        
         \midrule
         \bf{\texttt{CELLO}} (Ours)  &\textcolor{green}{\ding{52}} &\textcolor{green}{\ding{52}} &\textcolor{green}{\ding{52}} &\textcolor{green}{\ding{52}} &\textcolor{green}{\ding{52}} & Multi-choice  &\textcolor{green}{\ding{52}}&14,094   \\
         
    \bottomrule
    \end{tabular}
    \vspace{-2mm}
        \caption{
        Comparison of \texttt{CELLO} with existing causality-related vision-language datasets. Under the ``Question Types'' column, the abbreviations ``Disc.'', ``Assoc.'', ``Interv.'', and  ``CF.'' represent the four causal levels: Discovery, Association, Intervention, and Counterfactual, respectively. 
        ``\textcolor{red}{\ding{55}}'' denotes the absence of causality, ``\textcolor{yellow}{\ding{51}}'' denotes commonsense causality, and ``\textcolor{green}{\ding{52}}'' denotes both commonsense and formal causality (with causal graph).}
    \label{tab:benchmark}
    \vspace{-4mm}
\end{table*}

Previous work has primarily focused on commonsense causality between events and/or actions in a vision-language context~\cite{zellers2019vcr, park2020visualcomet, kim-etal-2022-cosim}, but often neglects the fine-grained causal relationships between humans and objects, between humans, and between objects. This limits the effectiveness of decision-making in real-world environments, such as embodied intelligent agents~\cite{cheong2024causal,gupta2024essential} and autonomous driving systems~\cite{ramanishka2018toward}. For instance, as illustrated in Figure~\ref{fig:example}, a model might respond ``\emph{yes}'' to the request, ``\emph{A child needs to reach something high. Can you move this chair for her to use?}'' This response overlooks the critical human-object causal relationship that ``\emph{the chair supports an old man}''\footnote{The human-object causal relationship will be detailed in Section \ref{sec:definition}.}, which would lead to a more reasonable decision. Moreover, these studies typically do not explicitly define the underlying causal graphs for key entities, rendering it challenging to systematically investigate the formal causal reasoning ability of LVLMs.

To address this, we first introduce a fine-grained and unified definition of causality in the vision-language context, drawing inspiration from the concept of \emph{causal dispositions} \citep{mumford2011getting,lopez2017discovering}.
We define a causal relationship as existing when one entity inherently possesses the ability to influence the state of another entity. This relationship can be further clarified through counterfactual reasoning \citep{pearl2009causality,peters2017elements}: if the ``\emph{cause}'' entity were absent, the ``\emph{effect}'' entity would not sustain its current state.
This includes interactions such as ``support'' and ``hold'', as well as spatial positioning between \emph{humans and humans}, \emph{humans and objects}, and \emph{objects and objects}. Using this foundational definition, we extract corresponding causal graphs from scene graphs in existing vision-language datasets and formulate questions based on these graph types. This results in \texttt{CELLO}, a novel dataset consisting of 14,094 causal questions across all four causal rungs of the \emph{Ladder of Causation}\footnotemark[2]~\cite{pearl2018book, bareinboim2022pearl, chen2024causal}: \emph{discovery}, \emph{association}, \emph{intervention}, and \emph{counterfactual}. As summarized in Table~\ref{tab:benchmark}, these questions cover various scenarios requiring different levels of causal reasoning abilities, allowing \texttt{CELLO} to offer a more comprehensive assessment of formal causal reasoning in LVLMs compared to other datasets.

\footnotetext[1]{In these work, not all questions are related to causality. We selectively extract those questions that are causality-related by filtering based on question type, and then tally the counts of these filtered instances.}
\footnotetext[2]{Following the extension by~\citet{chen2024causal}, we include \emph{(causal) discovery} into the ladder of causation. Please also refer to Section~\ref{sec:pre}.}

To elicit causal reasoning in LVLMs, we propose \texttt{CELLO-CoT}, a causally inspired chain-of-thought prompting strategy~\cite{wei2022chain,jin2023cladder,chen2024quantifying}. \texttt{CELLO-CoT} prompts LVLMs to systematically extract key entities, identify corresponding causal graphs, determine task types, and compile relevant causal knowledge to generate informed responses, enabling them to tackle challenging causal tasks in \texttt{CELLO}.

Through extensive experiments on \texttt{CELLO} with several leading LVLMs, we have observed several key findings:
1) Existing LVLMs perform poorly on causal reasoning tasks, with some models (e.g., BLIP-2~\cite{li2023blip} and Claude-3-sonnet~\cite{anthropic2024claude}) even underperforming random guessing, indicating substantial room for improvement.
2) There is notable variability in how different models perform across various types of causal reasoning tasks, reflecting distinct strengths and weaknesses of each model.
3) The \texttt{CELLO-CoT} strategy significantly enhances the performance of LVLMs on causal tasks, exemplified by an 11\% accuracy increase in GPT-4o. 
4) Robustness testing indicates that LVLMs' understanding of causal relationships is vulnerable, e.g.,  the performance of Qwen-VL~\cite{bai2023qwen} significantly drops from 49\% to 4\%.

Overall, our main contributions are as follows: 
\begin{itemize}[leftmargin=*, itemsep=0pt, parsep=0pt]
\item We introduce a fine-grained and unified definition of causality in the vision-language context, extending beyond the traditional focus on commonsense causality.
\item We construct \texttt{CELLO}, a novel dataset designed to rigorously evaluate the causal reasoning abilities of LVLMs. This dataset consists of 14,094 causal questions spanning all four causal levels, offering a comprehensive benchmark for assessment.
\item We propose \texttt{CELLO-CoT}, a causally inspired chain-of-thought prompting strategy, to effectively elicit the causal reasoning in LVLMs.
\item We conduct extensive experiments on ten leading LVLMs to assess their performance on causal reasoning tasks. Our analysis identifies their specific limitations and provides valuable insights for future research.
\end{itemize}
\section{Preliminaries}
\label{sec:pre}
\subsection{The Ladder of Causation} 
Causation refers to the cause-and-effect relationship where a change in one variable (the \emph{cause}) leads to a change in another (the \emph{effect}).
\textit{The Ladder of Causation}, proposed by \citet{pearl2018book}, builds a structured framework to illustrate the hierarchy of causal reasoning tasks, including  \emph{Association (Rung 1)}, \emph{Intervention (Rung 2)}, and \emph{Counterfactual (Rung 3)}. Following the extension by \citet{chen2024causal}, we incorporate \emph{(Causal) Discovery (Rung 0)} into this framework, establishing a more comprehensive four-rung ladder.

\paragraph{Rung 0: Discovery.} Causal discovery involves identifying cause-effect pairs from observational data, without prior knowledge of the underlying causal relationships. This fundamental step is crucial for establishing the initial causal structure within a given context~\cite{spirtes2001causation,peters2017elements,glymour2019review,zanga2022survey}. For example, ``\emph{Is there a causal relationship between talent and famous?}''

\paragraph{Rung 1: Association.} This rung focuses on identifying potential dependencies between variables, such as conditional relationships.
These dependencies can be effectively modeled by using Bayesian Networks~\cite{pearl1988probabilistic, goertzel2008probabilistic}, which represent a set of variables via a directed acyclic graph (DAG). For instance, ``\emph{How often do I become famous when I have talent?}''

\paragraph{Rung 2: Intervention.} This level goes beyond mere observation to explore the effects of manipulating certain variables. 
For instance, ``\emph{What if I have talent, will I become famous?}" By using the $do$-operator ~\cite{pearl1995causal}, we can model the effects of specific actions and determine their causal influence on other variables.

\paragraph{Rung 3: Counterfactual.} Counterfactual considers hypothetical scenarios to understand what could have happened under different conditions.
For instance, one might ask, ``\emph{If I have not gotten any talent, would I be famous?}''.

\subsection{Causal Graphical Models}
Causal graphical models (or causal models), utilize DAGs, referred to as causal graphs, to depict and analyze causal relationships between variables. In these models, nodes represent variables, and edges indicate direct causal influences \citep{pearl2009causality,peters2017elements}. These models are fundamental in understanding causal dynamics, predicting the effects of interventions, and addressing confounding across various disciplines such as epidemiology, economics, and psychology \citep{imbens2015causal,waldmann2017oxford}. 
Therefore, causal graphical models are crucial for elucidating complex causal relationships and facilitating decision-making processes in complex systems.
\section{Causality in Vision-Language Context}
\label{sec:definition}
\begin{figure}
\centering  
\includegraphics[width=0.42\textwidth]{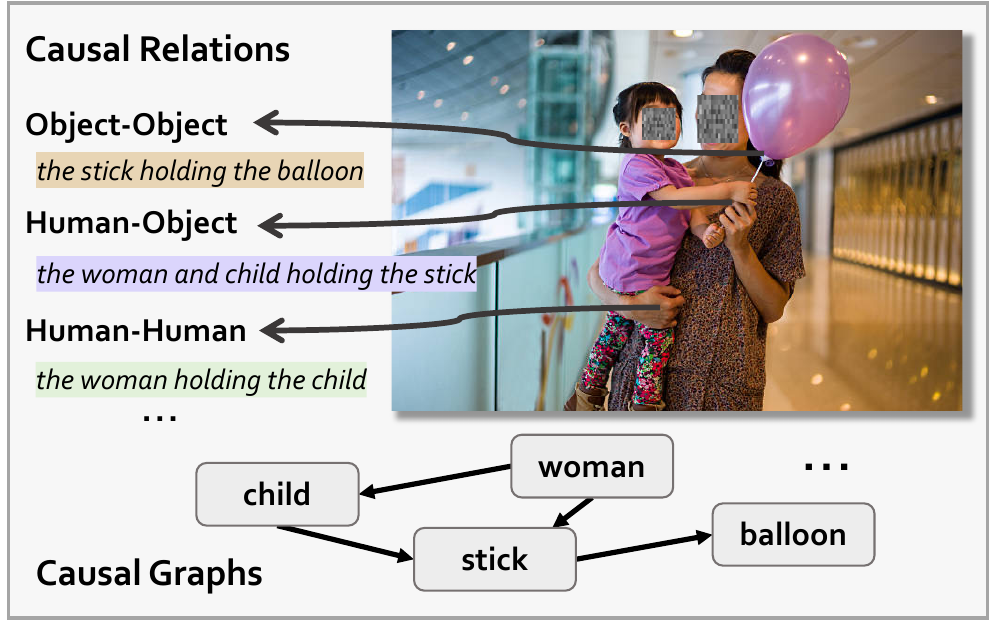}
\vspace{-1mm}
\caption{Three different causal relationships considered in the vision-language context: \emph{object-object}, \emph{human-object}, and \emph{human-human} causal relationships.}
\label{fig:causal}
\vspace{-4mm}
\end{figure}

We introduce a fine-grained and unified definition of causality in the vision-language context, inspired by the concept of \emph{causal dispositions}~\cite{mumford2011getting,lopez2017discovering}. We propose that a causal relation between entities exists when one entity influences the state of another. 
To be specific, a causal effect is present if one entity causes another to sustain its current state. This can be further explicated through counterfactual reasoning: if the ``\emph{cause}'' entity were absent, the ``\emph{effect}'' entity would not continue in its current state.
As shown in Figure~\ref{fig:causal}, 
we identify three distinct categories of causal relations in a scene: \emph{object-object}, \emph{human-object}, and \emph{human-human} causal relations.

\paragraph{Object-Object Causal Relation.}
This represents interactions between objects, such as ``\emph{the stick holding the balloon.}'' Without the stick, the balloon would not be in its current position (attached to the stick). Hence, the stick causes the balloon to maintain its current state. 
Identifying these relationships is crucial for understanding the physical interactions and dependencies within a scene. 

\paragraph{Human-Object Causal Relation.} This involves interactions between humans and objects, such as ``\emph{the woman and child holding the stick.}'' Without the woman and child, the stick would fall. Thus, both the woman and child cause the stick to sustain its current state. Recognizing these relations helps in comprehending human actions and their interactions with the surrounding environment.

\paragraph{Human-Human Causal Relation.} This denotes interactions between humans, such as ``\emph{the woman holding the child.}'' Without the woman, the child would not be held. Therefore, the woman causes the child to remain held. Understanding these relationships is essential for interpreting social interactions and human behaviors in a scene.

The causal graph depicted in Figure~\ref{fig:causal} shows how entities in the scene are interconnected via causal relationships. Understanding these causal relations facilitates more precise and significant interpretations of complex scenes. For example, in embodied artificial intelligence~\cite{gupta2024essential} and autonomous driving systems~\cite{ramanishka2018toward}, robots or vehicles should make decisions based on the causal relationships between entities within their environments.
\begin{figure*}
\centering  
\includegraphics[width=0.85\textwidth]{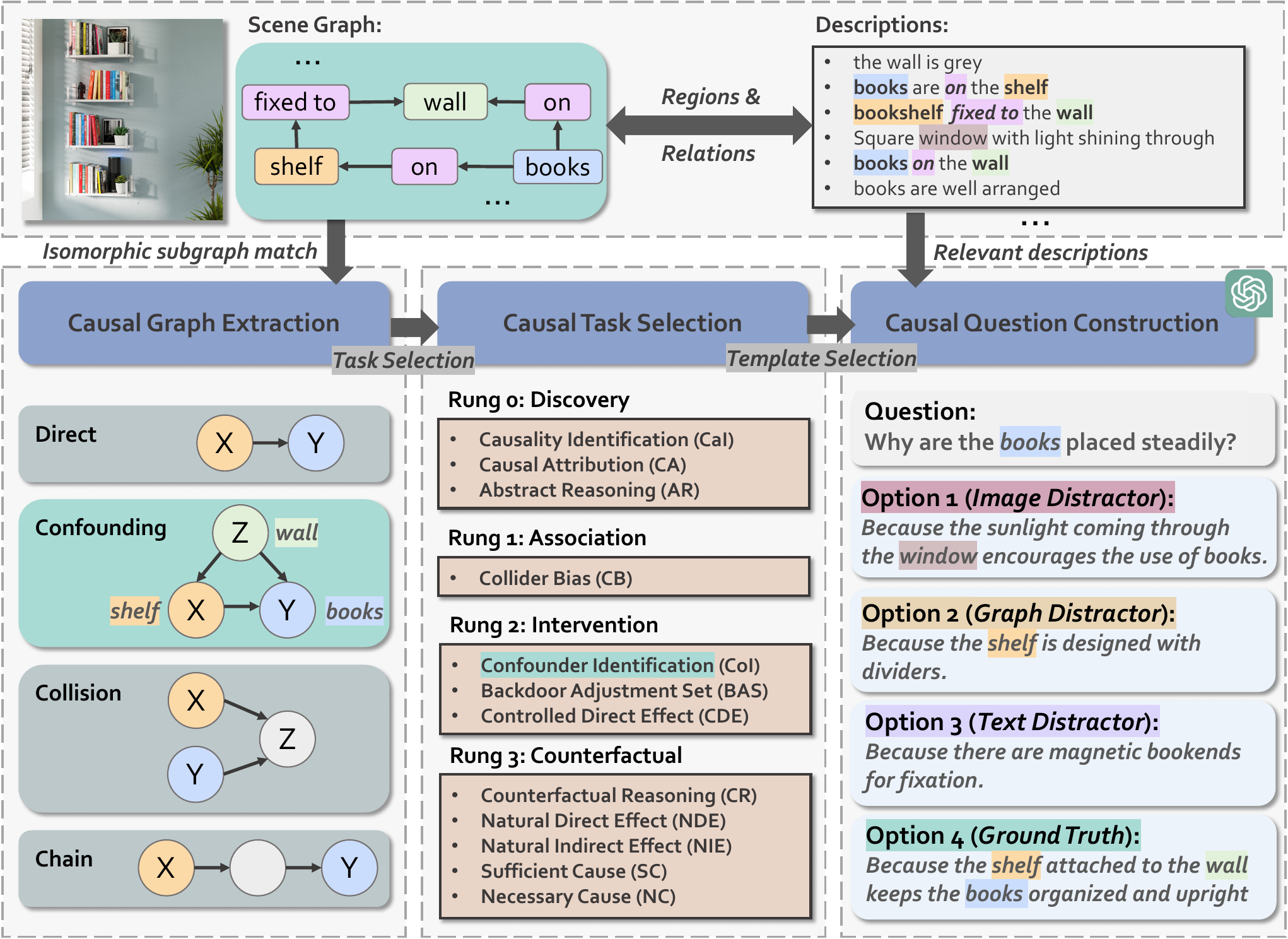}
\caption{Dataset construction pipeline of \texttt{CELLO} (using \emph{confounder identification} task as an example). First, we extract causal graphs from scene graphs that include relationships and regions within an image. Then, we select corresponding causal tasks based on the ladder of causation. Finally, causal questions are constructed by employing templates with an LLM. We consider four types of causal graphs and twelve different causal tasks in total.}
\label{fig:dataset}
\vspace{-2mm}
\end{figure*}
\section{The \texttt{CELLO} Dataset}
In this section, we elaborate on the dataset construction process based on the definition of causality as discussed in Section~\ref{sec:definition}. 
As shown in Figure~\ref{fig:dataset}, this process consists of three main steps: causal graph extraction, causal task selection, and causal question construction. 

\subsection{Causal Graph Extraction}
\label{subsec:cge}
The dataset construction begins with preprocessing the Visual Genome dataset~\cite{krishna2017visual}, utilizing its comprehensive suite of images along with corresponding scene graphs and descriptions.
From these resources, we construct causal graphs based on the relationships described between entities. Specifically, we first catalog and analyze every relationship type present in Visual Genome, with a focus on those signifying arrangement, positioning, and other significant interactions, such as those labeled ``support'', ``fixed to'', and ``hold''. 
Then, we compile a set of graph templates drawn from multiple sources in the literature~\cite{pearl2018book, bareinboim2022pearl, jin2023cladder, chen2024causal}, including \emph{direct}, \emph{confounding}, \emph{collision}, and \emph{chain}, as shown in Figure~\ref{fig:dataset}. These templates illustrate various toy problems in causal reasoning using well-defined graph structures. Finally, we perform isomorphic subgraph matching against these predefined templates to determine the type of causal graph extracted. For example, in Figure~\ref{fig:dataset}, the relationships extracted from the scene graph between ``wall'', ``shelf'', and ``books'' are matched to the ``confounding'' type of graph.

\subsection{Causal Task Selection}
To ensure comprehensive coverage, we select representative causal tasks of the ladder of causation from previous literature~\cite{pearl2018book, bareinboim2022pearl, jin2023cladder, chen2024causal}. For example in Figure~\ref{fig:dataset}, for the causal graph type of confounding, we could select the \emph{confounder identification} task. 
In total, we consider twelve distinct causal tasks as follows, and the mapping between causal graph types and causal tasks is presented in Table~\ref{tab:dataset_stat2} in the Appendix.
\paragraph{Discovery (Rung 0).} We include causal tasks such as \textbf{causality identification} (CaI, e.g., ``\emph{Which of the following elements is crucial for the girl's safety?}''), \textbf{causal attribution} (CA, e.g., ``\emph{What indirectly causes the balloon's stability?}''), and \textbf{abstract reasoning} (AR, e.g., ``\emph{What is indirectly influenced by the wave's force?}''). 
\paragraph{Association (Rung 1).} We consider \textbf{collider bias} (CB, e.g., ``\emph{Why don't the balloons fly away?}'').
\paragraph{Intervention (Rung 2).} We inquire about \textbf{confounder identification} (CoI, e.g., ``\emph{Why are the books placed steadily?}''), \textbf{backdoor adjustment set} (BAS, e.g., ``\emph{To assess the relationship between the solidity of shelves and the stability of books, which of the following variables should we control for?}
''), and \textbf{controlled direct effect} (CDE, e.g., ``\emph{If the state of the wall is not changed and the shelves become unstable, will the books drop?}''). 
\paragraph{Counterfactual (Rung 3).} We explore counterfactual scenarios such as \textbf{counterfactual reasoning} (CR, e.g., ``\emph{If the shelf has fallen down, would the books still be placed steadily?}''), \textbf{natural direct effect} (NDE, e.g., ``\emph{If the retainer of the shelf has been removed, would the books drop?}''), \textbf{natural indirect effect} (NIE, e.g., ``\emph{If the shelf has been fixed to a unstable wall, would the books stay steady?}''), \textbf{sufficient cause} (SC, e.g., ``\emph{If the wall has fallen down, would the books drop?}''), and \textbf{necessary cause} (NC, e.g., ``\emph{If the balloons has flown away, would the woman let go?}'').

\subsection{Causal Question Construction}
\label{subsec:cqc}
For question construction, we design templates for each task type in advance, with examples available in Appendix~\ref{app:prompt_question}. Each template includes a detailed task instruction along with several easily comprehensible demonstrations. The demonstration provides: 1) \emph{Relevant descriptions}, which are extracted from the dataset descriptions that are associated with the core entities. For instance, ``\emph{books are on the shelf}'', as shown in Figure~\ref{fig:dataset}. 2) \emph{Causal graph}, which is constructed through the process of Section~\ref{subsec:cge}. Each edge of the graph is expressed in textual form, such as ``\emph{shelf supports books}''. 3) \emph{Constraints}, which ensure the validity of the question and prevent information leakage, such as ``\emph{do not include `shelf' or `wall' in your generated question}''.
Using the template, an LLM (e.g., ChatGPT) is prompted to generate causal questions by applying in-context learning~\cite{brown2020language}. 

As for answer construction, we employ two settings. The first is a multiple-choice format, consisting of the correct answer and three distractors. The correct answer is derived by applying causal reasoning rules. For instance, in Figure~\ref{fig:dataset}, the ``wall'' is a \emph{confounder} because it affects both the stability of the ``shelf'' and the placement of the ``books''. Hence, the correct answer should include both ``shelf'' and ``wall''. The three distractors are constructed using the entities based on the following constraints:
1) \emph{Irrelevant entities} (\textbf{Image Distractor}): These entities are present in the image but absent from the causal graph, such as ``window''.
2) \emph{Partially correct entities} (\textbf{Graph Distractor}): These entities are present in the causal graph but only represent part of the correct answer, such as ``shelf''.
3) \emph{Induced entities} (\textbf{Text Distractor}): These entities are neither in the image nor in the causal graph but introduced solely from the question text, such as ``bookends''. This distractor can also be seen as a \emph{object hallucination}~\cite{parcalabescu-etal-2022-valse, lovenia2023negative} or \emph{language bias}~\cite{abbasnejad2020counterfactual, chen2024quantifying}. 
The correct answers and distractors can be further refined by an LLM to ensure natural and diverse expression. 
Additionally, for certain tasks, we also provide binary questions, where responses are limited to ``yes'' or ``no'', maintaining a nearly equal distribution between the two.
\subsection{Dataset Statistics and Quality Analysis}
\label{subsec:stat} 
\paragraph{Statistics of Four Rungs.}
Following the dataset construction process above, we randomly select appropriate images from the Visual Genome dataset to extract the corresponding causal graphs and then to generate causal questions. The statistical data for the 12 causal tasks across four causal rungs is detailed in Appendix~\ref{app:data}.

\paragraph{Question Quality.}
\begin{figure}
\centering  
\includegraphics[width=0.48\textwidth]{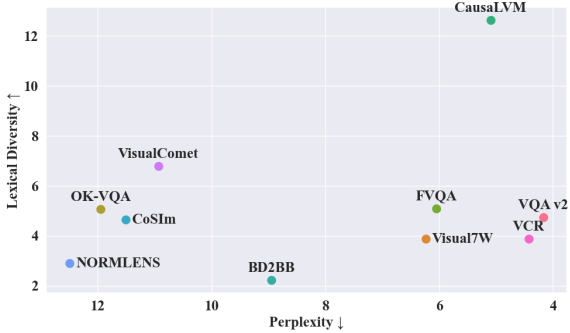}
\caption{Question quality of \texttt{CELLO} compared to other vision-language datasets in terms of lexical diversity and fluency.}
\label{fig:question_quality}
\end{figure}
We analyze the lexical diversity and fluency of the generated questions, with baselines and metrics detailed in Appendix~\ref{app:question_quality}. From Figure~\ref{fig:error} (a), \texttt{CELLO} shows superiority in lexical diversity and fluency.

\paragraph{Human Evaluation.}
We also conduct a human evaluation to validate the quality of the generated questions. Results in Appendix~\ref{app:human_evaluation} show that 91.7\% of questions are deemed valid by annotators, further demonstrating the quality of our datasets.
\section{The \texttt{CELLO-CoT} Strategy}
\begin{figure}
\centering  
\includegraphics[width=0.45\textwidth]{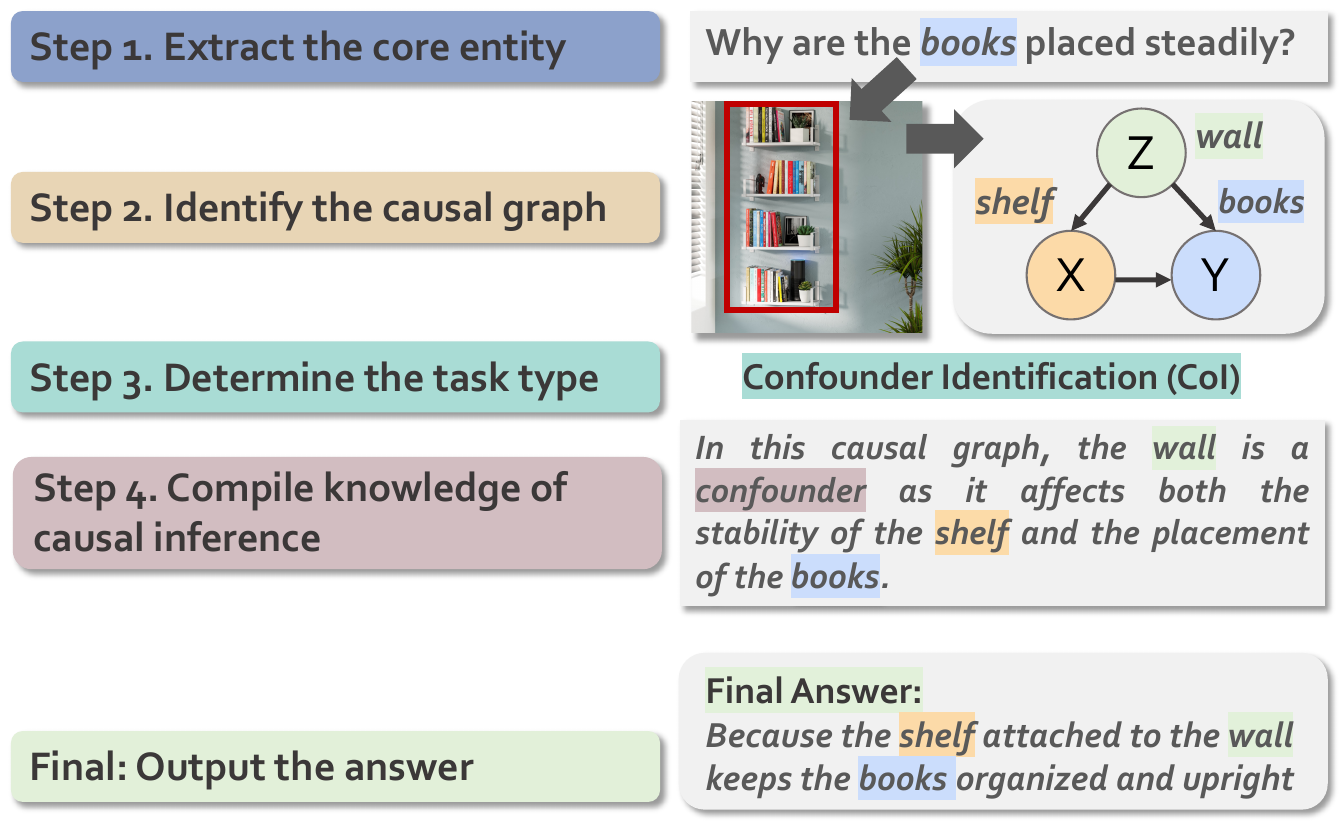}
\caption{Illustration of our \texttt{CELLO-CoT} strategy.}
 \vspace{-4mm}
\label{fig:cot}
\end{figure}

To enhance the capability of LVLMs in accurately responding to the questions in \texttt{CELLO}, 
we propose \texttt{CELLO-CoT}, a causally inspired chain-of-thought prompting strategy. It decomposes each causal question into multiple clear and manageable steps, enabling a sequentially structured analysis that supports effective problem-solving.

Given a causal question \( q \) with a corresponding image $i$, we provide LVLMs with a series of instructions \( \ell := (s_1, \ldots, s_4) \), including detailed descriptions of the four steps \( s_1, \ldots, s_4 \) depicted in Figure~\ref{fig:cot}. 
This structured approach includes 1) extracting core entities from the question text; 2) identifying the causal graph structure represented in the image; 3) determining the type of causal task, and 4) compiling knowledge of causal inference relevant to the current task (e.g., the core concepts about ``confounder'' in Figure~\ref{fig:cot}). 
The model $f_{\mathrm{LVLMs}}: \boldsymbol{s}_i \mapsto \boldsymbol{r}_i$ then autoregressively generates responses \( r_1, \ldots, r_4 \) corresponding to these steps. The final answer output will consider all these reasoning processes.
Compared to the standard strategy of directly posing questions to models, \texttt{CELLO-CoT} imposes an \emph{inductive bias}~\cite{jin2023cladder} on LVLMs, providing an effective solution to tackle causal reasoning problems.

\section{Experiments}
\begin{table*}
    \renewcommand
    \arraystretch{1.0}
    \centering
    \small
    \setlength{\tabcolsep}{2pt}
    \begin{tabular}{l|cccc|c|cccc|cccccc|c|c|c}
    \toprule
         \multirow{2}{*}{\bf{Model}} & \multicolumn{4}{c|}{\textbf{Discovery}} & \multicolumn{1}{c|}{\textbf{Assoc.} } & \multicolumn{4}{c|}{\textbf{Intervention} } &
         \multicolumn{6}{c|}{\textbf{Counterfactual} }&
         \multirow{2}{*}{\textbf{BIN.} }&
         \multirow{2}{*}{\textbf{MCQ.} }&
         \multirow{2}{*}{\textbf{ALL.} } \\ 
         \cmidrule(lr){2-5}\cmidrule(lr){6-6}\cmidrule(l){7-10} \cmidrule(l){11-16}
         & CaI   & CA   & AR &Avg. & CB  & CoI  & BAS &CDE &Avg. &CR & NDE &NIE &SC & NC &Avg. & & &\\ \midrule
          Random &0.25 &0.25 &0.25 &0.25 &0.25 &0.25&0.25 &0.50 &0.33 &0.50 &0.50 &0.50 &0.50 &0.50 &0.50 &0.50 &0.25 &0.38 \\ \midrule
         BLIP-2 &0.32 &0.26 &0.31 &0.30 &0.25 &0.30 &0.16  &0.51 &0.32 &0.57 &0.53 &0.45 &0.41 &0.44 &0.48 &\sethlcolor{mention}\hl{0.49} &0.27 &0.38\\
        LLaVA-M  &0.56 &0.54 &0.28 &0.46 &0.45 &0.37 &0.60 &0.38 &0.45 &0.58 &0.64 &0.76 &0.82 &\sethlcolor{second}\hl{0.77} &0.71 &0.66 &0.47 &0.56\\
         LLaVA-V  &0.52 &0.51 &0.35 &0.46 &\sethlcolor{mention}\hl{0.54} &\sethlcolor{second}\hl{0.51} &0.58  &\sethlcolor{mention}\hl{0.33} &0.47 &0.58 &0.67 &0.71 &0.85 &0.35 &0.63 &0.58 &0.50 &0.54\\
         BakLlava  &0.49 &0.52 &0.32 &0.44 &0.43 &0.38 &\sethlcolor{second}\hl{0.63}  &0.37 &0.46 &0.52 &0.63 &0.74 &0.89 &\sethlcolor{best}\hl{0.87} &\sethlcolor{second}\hl{0.73} &0.67 &0.46 &0.57\\
         MiniCPM &0.49 &0.45 &0.23 &0.39 &\sethlcolor{best}\hl{0.61} &0.50 &0.48 &0.59 &0.52 &0.63  &0.69 &0.59 &0.87  &0.53 &0.66 &0.65 &0.46 &0.56\\
         Qwen-VL &0.42 &0.51 &0.33 &0.42 &\sethlcolor{second}\hl{0.55} &\sethlcolor{best}\hl{0.55} &0.54 &0.55 &\sethlcolor{second}\hl{0.55} &0.54 &0.59 &0.58 &0.42 &0.35 &0.50 &0.51 &0.48 &0.49\\
         \midrule
         Claude-3-sonnet &0.33 &0.34 &0.19 &0.29 &0.38 &0.32 &0.27 &0.35 &0.31 &0.56  &0.52 &0.51 &0.77 &0.28 &0.53 &\sethlcolor{mention}\hl{0.49} &0.30 &0.40\\
         Claude-3-opus&0.54 &0.50 &0.35 &0.46 &0.44 &0.39 &0.42  &0.51 &0.44 &0.55 &0.63 &0.63 &\sethlcolor{best}\hl{0.95} &0.30 &0.61 &0.59 &0.44 &0.52\\
         \midrule
         Gemini-1.5-Pro &0.56 &0.56 &0.34 &0.49 &0.32 &0.28 &0.43 &0.70 &0.47 &0.67 &0.70 &0.70 &0.80 &0.38 &0.65 &0.66 &0.41 &0.54 \\
         + \texttt{CELLO-CoT} &\sethlcolor{second}\hl{0.76} &\sethlcolor{second}\hl{0.68} &\sethlcolor{best}\hl{0.54} &\sethlcolor{second}\hl{0.66} &0.43 &0.32 &0.62 &\sethlcolor{second}\hl{0.71} &\sethlcolor{second}\hl{0.55} &\sethlcolor{best}\hl{0.74} &0.75 &0.73 &0.87 &0.46 & 0.71 &0.71 &\sethlcolor{second}\hl{0.56} &\sethlcolor{second}\hl{0.64} \\ \midrule
         GPT-4o &0.63 &0.57 &0.32 &0.51 &0.43 &0.29 &0.49 &\sethlcolor{second}\hl{0.71} &0.50 &0.66  &\sethlcolor{second}\hl{0.77 }&\sethlcolor{best}\hl{0.77} &0.83 &0.61 &\sethlcolor{second}\hl{0.73} &\sethlcolor{second}\hl{0.73} &0.45 &\sethlcolor{mention}\hl{0.59}\\
         + \texttt{CELLO-CoT} &\sethlcolor{best}\hl{0.83} &\sethlcolor{best}\hl{0.70} &\sethlcolor{second}\hl{0.52} &\sethlcolor{best}\hl{0.68} &0.50 &\sethlcolor{mention}\hl{0.35} &\sethlcolor{best}\hl{0.75} &\sethlcolor{best}\hl{0.81} &\sethlcolor{best}\hl{0.64} &\sethlcolor{second}\hl{0.72} &\sethlcolor{best}\hl{0.79} &\sethlcolor{best}\hl{0.77} &\sethlcolor{second}\hl{0.90} &0.61 & \sethlcolor{best}\hl{0.76} &\sethlcolor{best}\hl{0.76} &\sethlcolor{best}\hl{0.59} &\sethlcolor{best}\hl{0.70} \\ 
    \bottomrule
    \end{tabular}
    \vspace{-2mm}
        \caption{LVLMs' results on \texttt{CELLO}.  We report the standard accuracy for each causal task. ``Assoc.'' denotes Association, ``BIN.'' denotes binary questions, ``MCQ.'' denotes multiple-choice questions, and ``ALL.'' denotes all questions. The \sethlcolor{best}\hl{best} and  \sethlcolor{second}\hl{second-best} results, as well as the \sethlcolor{mention}\hl{mentioned} results in the main text are highlighted.}
    \label{tab:main}
        \vspace{-4mm}
\end{table*}

\subsection{Experimental Setup}
\paragraph{Datasets.}
We compose a test set consisting of 1,200 samples, distributed equally across 12 causal tasks in \texttt{CELLO}, with each task featuring 100 randomly selected samples. 

\paragraph{Baselines.}
We evaluate ten leading LVLMs in a zero-shot fashion, including four \emph{limited-access} LVLMs: Claude-3-sonnet, Claude-3-opus~\cite{anthropic2024claude}, Gemini-1.5-Pro~\cite{team2023gemini}, and GPT-4o~\cite{gpt2023openai},
and six \emph{open-source} LVLMs: BLIP-2 (6.7B) ~\cite{li2023blip}, LLaVA-Mistral (7B), BakLlava (7B), LLaVA-Vicuna (13B)~\cite{liu2023visual}, MiniCPM-Llama3-V-2.5 (8B)~\cite{xu2024llava-uhd}, and Qwen-VL(7B)~\cite{bai2023qwen}. Details on these models are provided in Appendix~\ref{app:baselines}. For consistent evaluation, we use standard accuracy metrics for all the models and tasks. Performance is also benchmarked against a random baseline (i.e., 0.5 for binary and 0.25 for multiple-choice questions).

\subsection{Main Results}
The evaluation results of LVLMs on \texttt{CELLO} are presented in Table~\ref{tab:main} and further illustrated with case studies in Appendix~\ref{app:case}. 

\paragraph{Overall Performance.}
1) Among all the LVLMs (without \texttt{CELLO-CoT}), GPT-4o achieves the highest overall accuracy (\sethlcolor{mention}\hl{0.59}), demonstrating superior performance across all task categories.
2) BLIP-2 and Claude-3-sonnet perform relatively poorly across all tasks. Notably, their scores on binary questions (\sethlcolor{mention}\hl{0.49}) fail to surpass the random baseline of 0.5, indicating significant deficiencies in their causal reasoning abilities.
3)  All models exceed the random baseline (0.25) on multiple-choice questions. However, no models (without \texttt{CELLO-CoT}) achieve a performance higher than 0.5, highlighting their inherent limitations.
4) Implementing our proposed \texttt{CELLO-CoT} strategy significantly enhances the performance of GPT-4o and Gemini-1.5-Pro across various causal reasoning tasks, thus confirming the effectiveness of our approach.

\paragraph{Ladder-Specific Results.}
1) \emph{Discovery Tasks}: GPT-4o (with \texttt{CELLO-CoT}) achieves the highest accuracy for discovery tasks (\sethlcolor{best}\hl{0.68}), notably in causality identification (\sethlcolor{best}\hl{0.83}) and causal attribution (\sethlcolor{best}\hl{0.70}).
2) \emph{Association Tasks}: MiniCPM-Llama3-V-2.5 (8B) leads for association tasks (\sethlcolor{best}\hl{0.61}), surpassing even higher-parameter models like LLaVA-Vicuna (13B, \sethlcolor{mention}\hl{0.54}). This indicates its superior handling of collider bias.
3) \emph{Intervention Tasks}: GPT-4o (with \texttt{CELLO-CoT}) excels in the controlled direct effect task (\sethlcolor{best}\hl{0.81}), but underperforms in the confounder identification task (\sethlcolor{mention}\hl{0.35}). Conversely, LLaVA-Vicuna performs poorly in the controlled direct effect task (\sethlcolor{mention}\hl{0.33}) but well in the confounder identification task (\sethlcolor{second}\hl{0.51}). These findings demonstrate variability in task-specific performance among different models.
4) \emph{Counterfactual Tasks}: GPT-4o (with \texttt{CELLO-CoT}) achieves high accuracy across all counterfactual tasks, particularly excelling in natural direct effect (\sethlcolor{best}\hl{0.79}) and natural indirect effect (\sethlcolor{best}\hl{0.77}).
This highlights its capacity for sophisticated counterfactual reasoning about hypothetical alternatives to actual conditions. Further details and illustrations of performance are available in Appendix~\ref{app:acc}.

\subsection{Ablation Study}
\begin{figure}
\centering  
\includegraphics[width=0.4\textwidth]{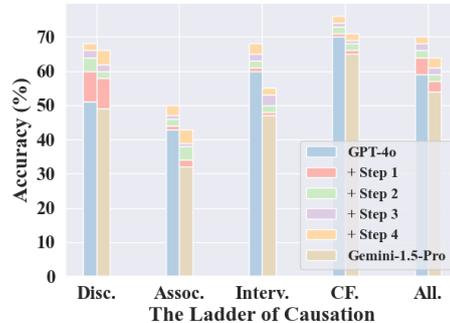}
\caption{Ablation study on our proposed \texttt{CELLO-CoT}, where ``Disc'' denotes discovery, ``Assoc'' denotes association, ``Interv'' denotes intervention, ``CF'' denotes counterfactual reasoning.}
\label{fig:ablation}
\end{figure}

We conduct ablation studies to evaluate the effect of each component in our \texttt{CELLO-CoT} prompting strategy, as shown in Figure~\ref{fig:ablation} (a):
1) Each step in the \texttt{CELLO-CoT} strategy contributes to performance gains at different rungs of the ladder of causation, demonstrating the effectiveness of our approach. 
2) Notably, \texttt{CELLO-CoT} yields more pronounced improvements in lower-level causal tasks (e.g., discovery), whereas its influence on higher-level causal tasks remains modest. This disparity suggests that more sophisticated strategies are necessary to address complex causal reasoning challenges.
3) For lower-level tasks like discovery, the primary factor is the extraction of core entities (Step 1). Conversely, for higher-level tasks, a deeper understanding of causal graphs and causal inference (Steps 2 to 4) becomes essential.

\subsection{Robustness Testing}
\label{subsec:rob}
\begin{figure}
\centering  
\includegraphics[width=0.4\textwidth]{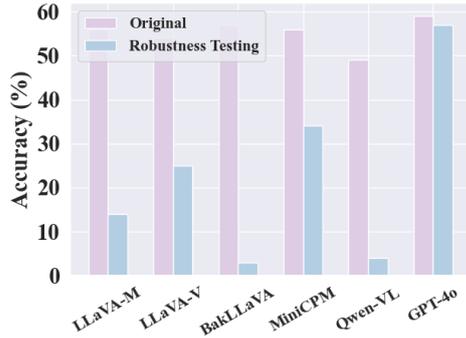}
\caption{Robustness testing across various LVLMs. It can be observed significant performance decline (e.g., BakLlava, from 0.57 to 0.03). }
\label{fig:robustness}
\end{figure}
We further conduct robustness tests on selected representative LVLMs. This involves reformulating the questions in the test set by incorporating additional premises and posing a plausible but contextually inappropriate request. The response options are limited to ``Yes'' and ``No'', with the correct answer consistently being ``No''. For example in the case of Figure ~\ref{fig:dataset}, the rephrased question could be, ``\emph{Bob needs support for his toys. Can you bring this shelf over?}" We implement this reformulation by using prompts with ChatGPT, detailed in the template provided in Appendix~\ref{app:prompt_rob}. 

From Figure~\ref{fig:ablation} (b), we observe that: 1) Faced with reformulated questions, LVLMs tend to respond affirmatively, focusing on the request's tone rather than the actual causal relationships depicted in the scene. For instance, in Figure~\ref{fig:dataset}, despite the shelf being occupied with the books, the models erroneously suggest bringing it over. This misalignment significantly diminishes the performance of these models, with notable declines seen in BakLlava and Qwen-VL, whose accuracies plummet from 0.57 and 0.49 to 0.03 and 0.04, respectively. 
2) GPT-4o, however, exhibits relatively stable performance. A closer examination of its responses reveals that it does not directly address the unreasonableness of the requests.
Instead, it typically responds with, ``\emph{No, I am a language model and cannot interact with the physical world.''} This response pattern likely results from its training, which included similar instructions during its development phase~\cite{ouyang2022training}. Further details of these findings are provided in Appendix~\ref{app:rob}.

\subsection{Error Analysis}
\begin{figure}
\centering  
\includegraphics[width=0.3\textwidth]{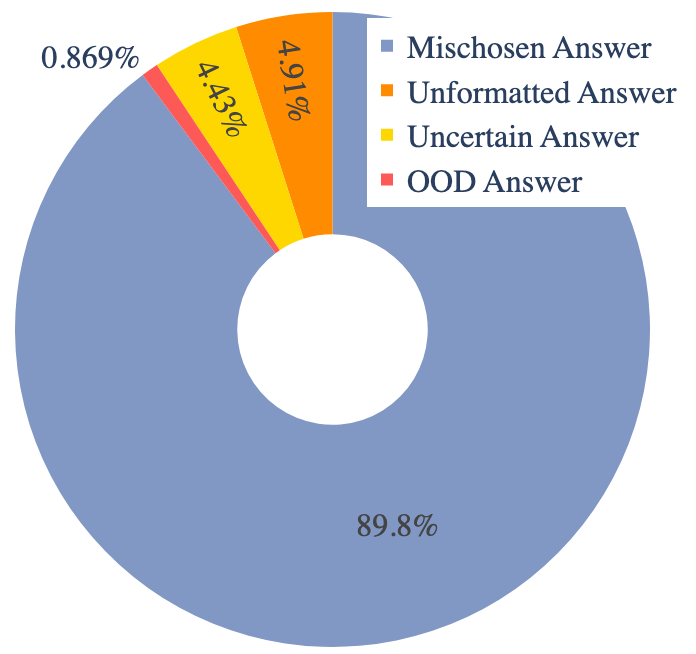}
\caption{Error Analysis of LVLMs.}
\label{fig:error}
\end{figure}
To understand why LVLMs struggle with \texttt{CELLO} deeply, we conduct a thorough error analysis. Figure~\ref{fig:error} (b) categorizes errors made by all models across 1200 test instances into four distinct types: 1) \textit{Mischosen Answer}: when models select an incorrect option, probably influenced by irrelevant visual or textual cues in the test instance. 2) \textit{Out-Of-Distribution (OOD) Answer}: when models provide an answer that is not among the given options, indicating a phenomenon often referred to as \emph{hallucination}~\cite{li2023halueval}. 3) \textit{Unformatted Answer}: where responses are incorrectly formatted and difficult to extract valid choices. 4) \textit{Uncertain Answer}: when models either explicitly state ``I don't know'' or demonstrate an inability to determine a definitive answer.
Detailed analyses focusing on models, tasks, ladder levels, and causal graph types can be found in Appendix~\ref{app:error}. Specific examples illustrating these error types are also provided in Appendix~\ref{app:case}.
\section{Related Work}
\paragraph{Causal Evaluation on Language Models.}
Several works have evaluated causality-related skills for NLP tasks. For example, \citet{sap2019atomic} investigate commonsense causality through "if-then" statements, while \citet{zhang2020reasoning} introduce reasoning tasks that consist of a series of steps towards a high-level goal. \citet{chen2022ergo} and \citet{chen2023cheer} focus on identifying cause-effect pairs to extract causal relations from document-level context.
With the increasing focus on LLMs and causality, numerous studies have aimed to evaluate the causal reasoning abilities of large language models (LLMs)~\cite{zhang2023understanding, kiciman2023causal, jin2023can, chen2023learning, zevcevic2023causal, jin2023cladder, chen2024causal}. Unlike these studies, our research focuses on causal relations within the vision-language context.

\paragraph{Large Vision-Language Models.}
Building on the success of LLMs, there has been growing research interest in large vision-language models (LVLMs) to enhance multimodal comprehension and generation~\cite{li2023blip, liu2023visual, xu2024llava-uhd, bai2023qwen, openai2023gpt4, anthropic2024claude}.
While previous assessments have noted deficiencies in LVLMs~\cite{fu2023mme, liu2023mmbench}, particularly in reasoning skills, their proficiency in understanding causal relationships remains less explored and requires further investigation.

\paragraph{Causality in Vision-Language Tasks.}
Early visual question answering (VQA) datasets like Visual7W~\cite{zhu2016visual7w} and VQA~\cite{goyal2017making} include some causality-related questions, typically beginning with ``\emph{Why}'' and focusing on specific events or actions. However, these questions are relatively simple and can be often answered even without consulting the images~\cite{abbasnejad2020counterfactual, zhu2020overcoming}. Subsequent datasets like FVQA~\cite{wang2017fvqa} and OKVQA~\cite{marino2019ok} aimed to elevate the complexity of questions by integrating external knowledge, but the presence of causality-related questions is notably sparse.
On the other hand, datasets such as VCR~\cite{zellers2019vcr} and VisualCOMET~\cite{park2020visualcomet}, derived from movie scenes, delve into the temporal dynamics of events and provide rationales for each query. Datasets like BD2BB~\cite{pezzelle-etal-2020-different}, COSIM~\cite{kim-etal-2022-cosim}, and NORMLENS~\cite{han2023reading} intervene on original questions in various scenarios. Nonetheless, they focus only on event-related commonsense causality, ignoring fined-grained interaction between humans and/or objects. Additionally, the absence of explicitly defined causal graphs means that the understanding of causality they foster is somewhat rudimentary. Our \texttt{CELLO} dataset (see Table \ref{tab:benchmark}) seeks to rectify these limitations by offering a thorough evaluation of causality, encompassing detailed interactions and explicit causal reasoning challenges.
\section{Conclusion}
In this paper, we introduce a fine-grained and unified definition of causality involving humans and objects. Building on the definition, we construct a novel dataset, \texttt{CELLO}, to assess the causal reasoning abilities of LVLMs. To elicit causal reasoning in LVLMs, we propose \texttt{CELLO-CoT}, a causally inspired chain-of-thought prompting strategy, enabling LVLMs to tackle challenging causal tasks in \texttt{CELLO}. Extensive experimental results, as well as further quantitative and qualitative analyses on \texttt{CELLO}, provide insights for future work.

\section*{Limitations}
Our dataset, \texttt{CELLO}, relies on the Visual Genome dataset \cite{krishna2017visual}, which is a large-scale visual language dataset featuring scene graphs and descriptions. Consequently, the quality of our dataset is inevitably influenced by the accuracy of the original annotations in Visual Genome. This includes challenges such as incorrect object identifications and unclear images.
Despite these issues, the quality analysis presented in Section~\ref{subsec:stat} demonstrates that the majority of questions are effectively constructed and valid.
Moreover, it is crucial to acknowledge that establishing causal relationships in real-world contexts often demands more intricate analyses, such as the examination of image sequences or video frames to discern the dynamics among recognized objects, actions, or scene changes. For example, in video analysis~\cite{lei2019tvqa+, yi2019clevrer, xiao2021next, li2022representation}, determining whether a person causes an object (e.g., a ball) to move involves a different set of reasoning skills. However, most current LVLMs are primarily designed for static image inputs, and enhancing their capabilities to handle dynamic visual data remains a vital area for future research.

\bibliography{custom}
\clearpage
\appendix
\section{Dataset Statistics}
\label{app:data}
\begin{table}
    \renewcommand
    \arraystretch{1.0}
    \centering
    \small
    \setlength{\tabcolsep}{2pt}
    \begin{tabular}{l|c|c}
    \toprule
         {\textbf{Dataset}}  & \textbf{\#{I, Q, A}} &\textbf{Len of Q / A} \\ \midrule
         \texttt{CELLO} &14,094 &14.9 / 6.9\\
         \ \ \ - Discovery &3,000  &11.7 / 1.1\\
         \ \ \ - Association &2,000  &7.98 / 14.9\\
          \ \ \ - Intervention &2,047  &13.9 / -\\
           \ \ \ - Counterfactual &7,047  &15.7 / -\\
         \midrule
         \multicolumn{3}{l}{\texttt{CELLO}-Discovery}  \\
         \ \ \ - Causality Identification (CaI) &1,000 &11.4 / 1.1 \\
        \ \ \ - Causal Attribution (CA) &1,000 &11.9 / 1.1\\
         \ \ \ - Abstract Reasoning (AR) &1,000  &11.8 / 1.1\\
         
         \multicolumn{3}{l}{\texttt{CELLO}-Association}  \\
        \ \ \ - Collider Bias (CB) &2,000 &7.98 / 14.9\\

        \multicolumn{3}{l}{\texttt{CELLO}-Intervention} \\
         \ \ \ - Confounder Identification (CoI) &349 &8.2 / 16.4\\
        \ \ \ - Backdoor Adjustment Set (BAS) &349 &25.3 / 1.1\\
        \ \ \ - Controlled Direct Effect (CDE) &1,349 &12.3 / -\\

         \multicolumn{3}{l}{\texttt{CELLO}-Counterfactual}  \\
        \ \ \ - Counterfactual Reasoning (CR) &2,000 &13.8 / -  \\
         \ \ \ - Natural Direct Effect (NDE) &1,349 &20.3 / -  \\
        \ \ \ - Natural Indirect Effect (NIE) &1,349 &12.3 / -  \\
         \ \ \ - Sufficient Cause (SC) &349 &15.6 / -  \\
         \ \ \ - Necessary Cause (NC) &2,000 &16.9 / -  \\
    \bottomrule
    \end{tabular}
       \caption{\label{tab:dataset_stat1} Dataset statistics of \texttt{CELLO} based on the ladder of causation. ``I, Q, A'' denotes images, questions, and answers, respectively. ``Len'' denotes length and ``-'' denotes binary questions where answers are limited to ``yes'' or ``no''.}
\end{table}

\begin{table}
    \renewcommand
    \arraystretch{1.0}
    \centering
    \small
    \setlength{\tabcolsep}{6pt}
    \begin{tabular}{l|c}
    \toprule
         {\textbf{Dataset}}  & \textbf{\#{I, Q, A}} \\ \midrule
         \texttt{CELLO} &14094 \\
         \ \ \ \ - direct &3000  \\
         \ \ \ \ - confounding &2094  \\
          \ \ \ \ - collision &4000 \\
           \ \ \ \ - chain &5000   \\
         \midrule
         \multicolumn{2}{l}{\texttt{CELLO}-direct}  \\
         \ \ \ \ - causality identification &1000   \\
         \ \ \ \ - counterfactual reasoning &2000  \\
         \multicolumn{2}{l}{\texttt{CELLO}-confounding}  \\
         \ \ \ \ - confounder identification &349  \\
         \ \ \ \ - backdoor adjustment set &349   \\
        \ \ \ \ - controlled direct effect &349   \\
         \ \ \ \ - natural direct effect &349   \\
        \ \ \ \ - natural indirect effect &349  \\
         \ \ \ \ - sufficient cause &349  \\
         \multicolumn{2}{l}{\texttt{CELLO}-collision} \\
         \ \ \ \ - collider bias &2000   \\
         \ \ \ \ - necessary cause &2000  \\
         \multicolumn{2}{l}{\texttt{CELLO}-chain}  \\
        \ \ \ \ - causal attribution &1000  \\
         \ \ \ \ - abstract reasoning &1000   \\
        \ \ \ \ - controlled direct effect &1000 \\
         \ \ \ \ - natural direct effect &1000   \\
         \ \ \ \ - natural indirect effect &1000   \\
    \bottomrule
    \end{tabular}
       \caption{\label{tab:dataset_stat2} Dataset statistics based on the type of causal graphs.}
\end{table}

In Table~\ref{tab:dataset_stat1}, we present data statistics of \texttt{CELLO} for the 12 causal tasks across four causal rungs.
For further insights, Table \ref{tab:dataset_stat2} provides data statistics by types of causal graphs.

\section{Quality Analysis Details}
\begin{figure}
\centering  
\includegraphics[width=0.4\textwidth]{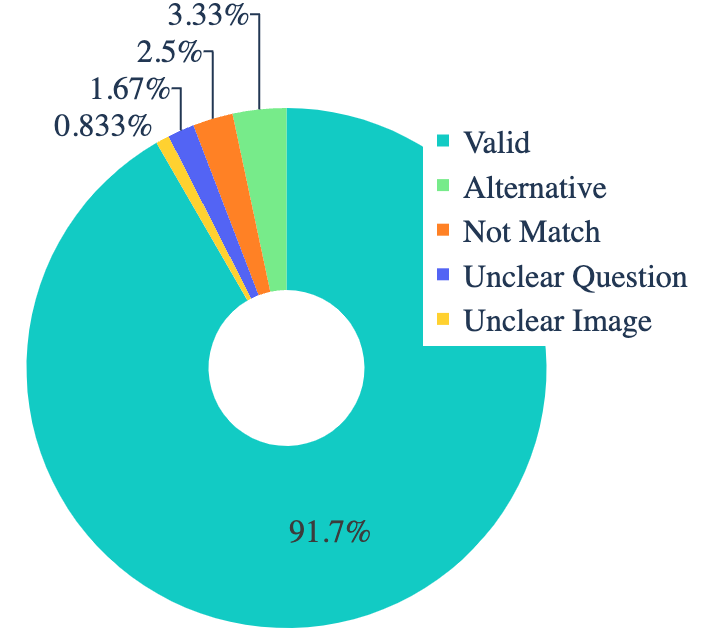}
\caption{Human evaluation results of \texttt{CELLO}.}
\label{fig:human_evaluation}
\end{figure}
\subsection{Question Quality}
\label{app:question_quality}
To ensure the quality of the comprising datasets, we analyze the lexical diversity and the fluency of the generated questions, which are useful for conducting a robust evaluation using questions that are linguistically diverse and coherent.
\paragraph{Baselines} We select extensive VQA datasets for comparison, including Visual7W~\cite{zhu2016visual7w}, VQA (v2)~\cite{goyal2017making},  FVQA~\cite{wang2017fvqa}, OK-VQA~\cite{marino2019ok}, VCR~\cite{zellers2019vcr}, VisualCOMET~\cite{park2020visualcomet}, BD2BB~\cite{pezzelle-etal-2020-different}, COSIM~\cite{kim-etal-2022-cosim} and NORMLENS~\cite{han2023reading}.
\paragraph{Evaluation Metrics}
For lexical diversity, following~\citet{chen2024quantifying}, we utilize three metrics that are not dependent on length: moving average type-token ratio (MATTR)~\cite{covington2010cutting}, measure of textual lexical diversity (MTLD)~\cite{mccarthy2005assessment}, and hypergeometric distribution diversity (HDD)~\cite{mccarthy2010mtld}. We average these three metrics for a unified assessment and employ the Lexical-Richness package~\cite{shen2022lexicalrichness} (version 0.5.03) for calculation. For fluency, we employ a pre-trained language model GPT2-large~\cite{radford2019language} with 774M parameters to compute the perplexity of the questions, which is often used as a measure by previous work~\cite{wang-etal-2019-paperrobot, cahyawijaya-etal-2021-indonlg}.

\begin{figure*}
\centering 
\includegraphics[width=1.0\textwidth]{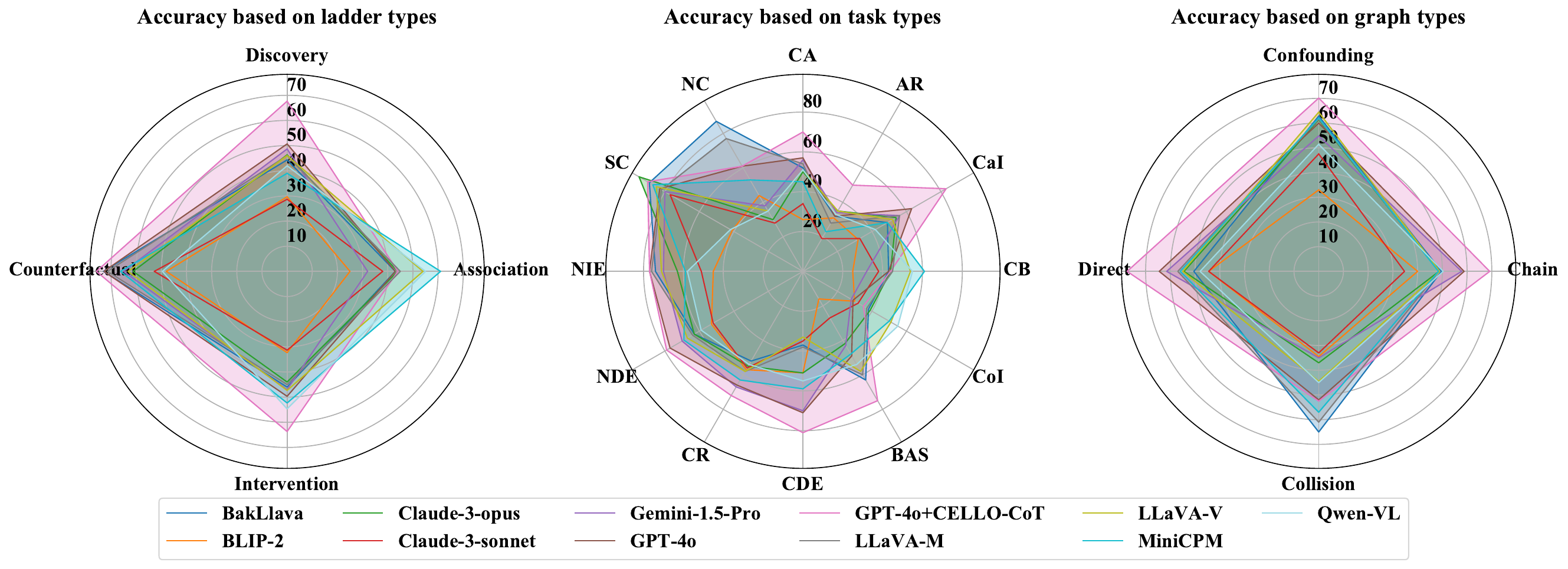}
\caption{Model results based on different ladder, task, and graph types, respectively.}
\label{fig:acc_main}
\end{figure*}

\begin{figure}
\centering  
\includegraphics[width=0.5\textwidth]{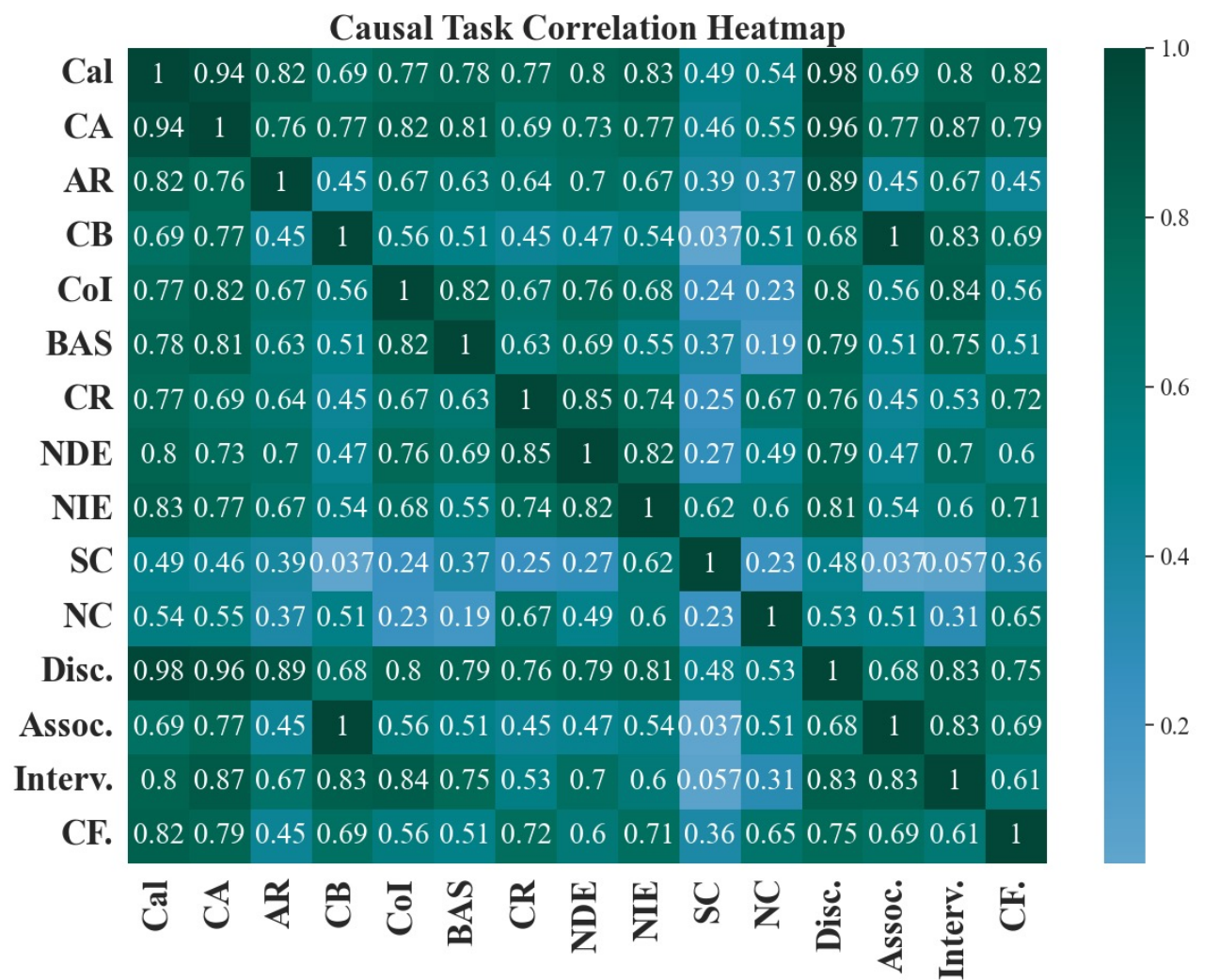}
\caption{Correlation among different causal tasks.}
\label{fig:correlation}
\vspace{-2mm}
\end{figure}

\subsection{Human Evaluation}
\label{app:human_evaluation}
\paragraph{Questions}
We conduct a human evaluation to validate and assess the quality of our \texttt{CELLO} dataset. We randomly sample 10 instances for each causal task, resulting in a total of 120 instances. 
The evaluation is conducted by two annotators independently, who are provided with detailed guidelines and illustrative examples before starting the evaluation process. For each question, given the image and ground truth answer, we first ask the annotators to determine whether: 1) the question is valid, 2) the question allows for an alternative answer, 3) the question does not match the ground truth, 4) the image is unclear, or 5) the question is unclear or ambiguous. The average inter-annotator agreement is 84.1\%~(Cohen's kappa).

As shown in Figure~\ref{fig:human_evaluation}, the results are encouraging, with 91.7\% questions being classified as valid by the annotators, further demonstrating the quality of our datasets.

\section{Baseline Details}
\label{app:baselines}
For open-source MLLMs, we consider the following baselines:

1) BLIP2~\cite{li2023blip}, which utilizes a scalable multimodal pre-training method to enable LLMs to understand images. We employ its BLIP2-OPT~\cite{zhang2022opt}-6.7B variant.

2) LLaVA~\cite{liu2023visual}, which translates images into texts of captions and bounding boxes, and prompts GPT-4 to generate a multimodal instruct-tuning dataset. We employ its three variants: LLaVA-Mistral (7B), BakLlava (7B), and LLaVA-Vicuna (13B).

3) Qwen-VL~\cite{bai2023qwen}, which builds upon Qwen~\cite{bai2023qwent} and employ 3-stage training pipeline.  Qwen-VL implements the grounding and text-reading ability by aligning image-caption-box tuples, i.e., it accepts image, text, and bounding box as inputs, and outputs text and bounding box. 

4) MiniCPM-Llama3-V-2.5~\cite{xu2024llava-uhd}, which is an end-side multimodal LLM designed for vision-language understanding, equipped with the OCR and instruction-following capability.

\section{Performance Details}
\label{app:acc}

As shown in Figure~\ref{fig:acc_main}, we visualize the model performance comparison based on different ladder types, task types, and graph types, respectively.

In Figure~\ref{fig:correlation}, we compute the Pearson correlation coefficients between LVLMs' results on different causal tasks and visualize the values in a heatmap. It can be seen that tasks within the same ladder exhibit higher correlation coefficients (e.g., the correlation coefficient between causal identification (CaI) and causal attribution (CA) is 0.94), whereas tasks between different ladders show relatively lower correlation coefficients.

\section{Robustness Testing Details}
\label{app:rob}
In Table~\ref{tab:rob}, we present the complete results of robustness testing. Since the rephrased questions differ from the original causal tasks, we report the answers based on the type of causal graphs.
\begin{table}
    \renewcommand
    \arraystretch{1.0}
    \centering
    \small
    \setlength{\tabcolsep}{6pt}
    \begin{tabular}{l|c|c|c|c|c}
    \toprule
         \bf{Model} &\bf{Dir.} &\bf{Conf.} & \bf{Coll.} &\bf{Ch.} &\bf{All.}\\ 
         \midrule
         Random &0.50 &0.50 &0.50 &0.50 &0.50 \\
         \midrule
        LLaVA-M &0.18 &0.12 &0.20 &0.13 &0.14 \\
         LLaVA-V &0.26 &0.24 &0.30 &0.24 &0.25 \\
         BakLlava &0.04 &0.03 &0.04 &0.02 &0.03 \\
         MiniCPM &0.31 &0.34 &0.36 &0.34 &0.34  \\
         Qwen-VL &0.06&0.03 &0.04 &0.03 &0.04\\
         \midrule
         GPT-4o &0.58 &0.57 &0.54 &0.58 &0.57\\
    \bottomrule
    \end{tabular}
    \vspace{-2mm}
        \caption{Robustness testing details based on different graph types. ”Dir.” denotes direct, “Conf.” denotes confounding, “Coll.” denotes collider, and “Ch.” denotes chain.}
    \label{tab:rob}
\end{table}

\section{Error Analysis Details}
\label{app:error}
\begin{figure*}
\centering  
\includegraphics[width=1\textwidth]{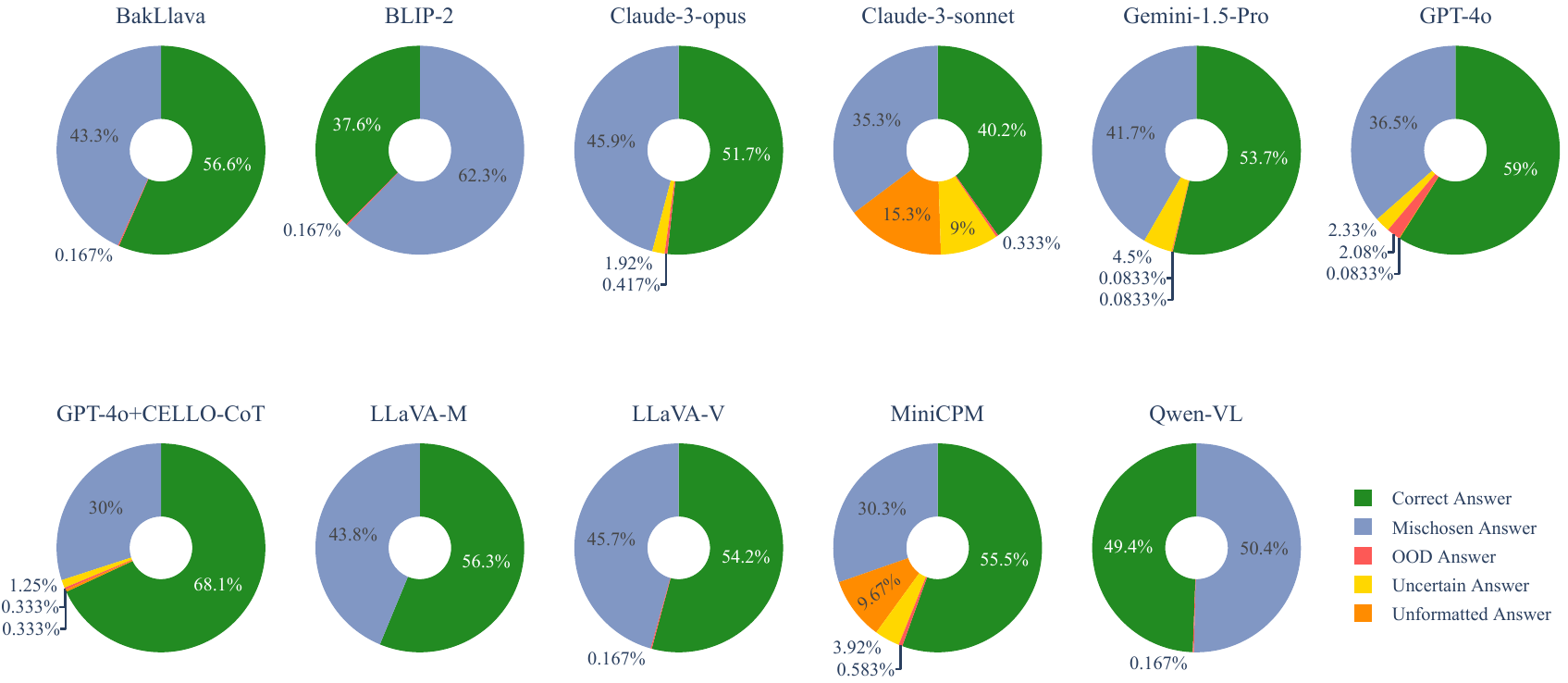}
\caption{Error analysis of models.}
\label{fig:model_error}
\vspace{-2mm}
\end{figure*}

\begin{figure*}
\centering 
\includegraphics[width=1\textwidth]{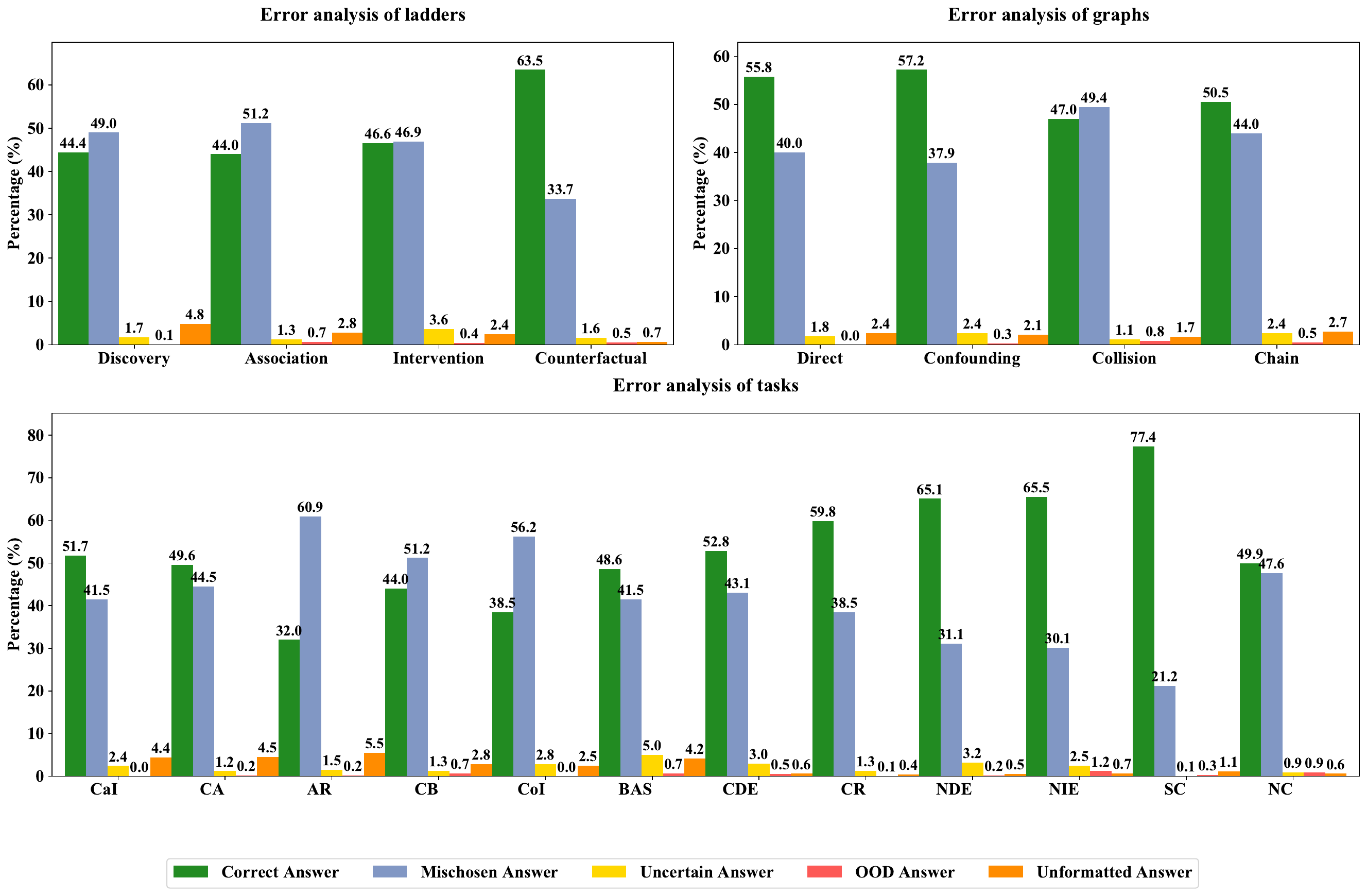}
\caption{Error analysis based on different ladder, graph, and task types.}
\label{fig:error_main}
\end{figure*}

We present a more detailed analysis of errors on models, ladders, causal graphs, and task types from Figure \ref{fig:model_error} to \ref{fig:error_main}, respectively. We include the proportion of correct answers for further comparisons.

Figure \ref{fig:model_error} shows the error distribution of different models on the test set. We also add the results of GPT-4o (w. \texttt{CELLO-CoT}). Among all the models, GPT-4o (w. \texttt{CELLO-CoT}) has the lowest proportion of errors. All kinds of error types that GPT-4o produces are reduced after applying \texttt{CELLO-CoT}. Moreover, it is noticeable that Claude-3-sonnet and MiniCPM-Llama3-V-2.5 have difficulty providing correctly formatted answers, leading to a relatively higher proportion of \textit{Unformatted Answer} types compared with other models.

From Figure \ref{fig:error_main}, we find that ladders and tasks with higher correctness tend to have less number of \textit{Uncertain Answer}s, \textit{OOD Answer}s, and \textit{Unformatted Answer}s. In contrast, the graph type with the highest correctness (i.e., \textit{Chain}) has a relatively higher proportion of \textit{Uncertain Answer}s.

\section{Prompt Templates}
\subsection{Question Generation}
\label{app:prompt_question}
\begin{figure*}
\centering  
\includegraphics[width=1.0\textwidth]{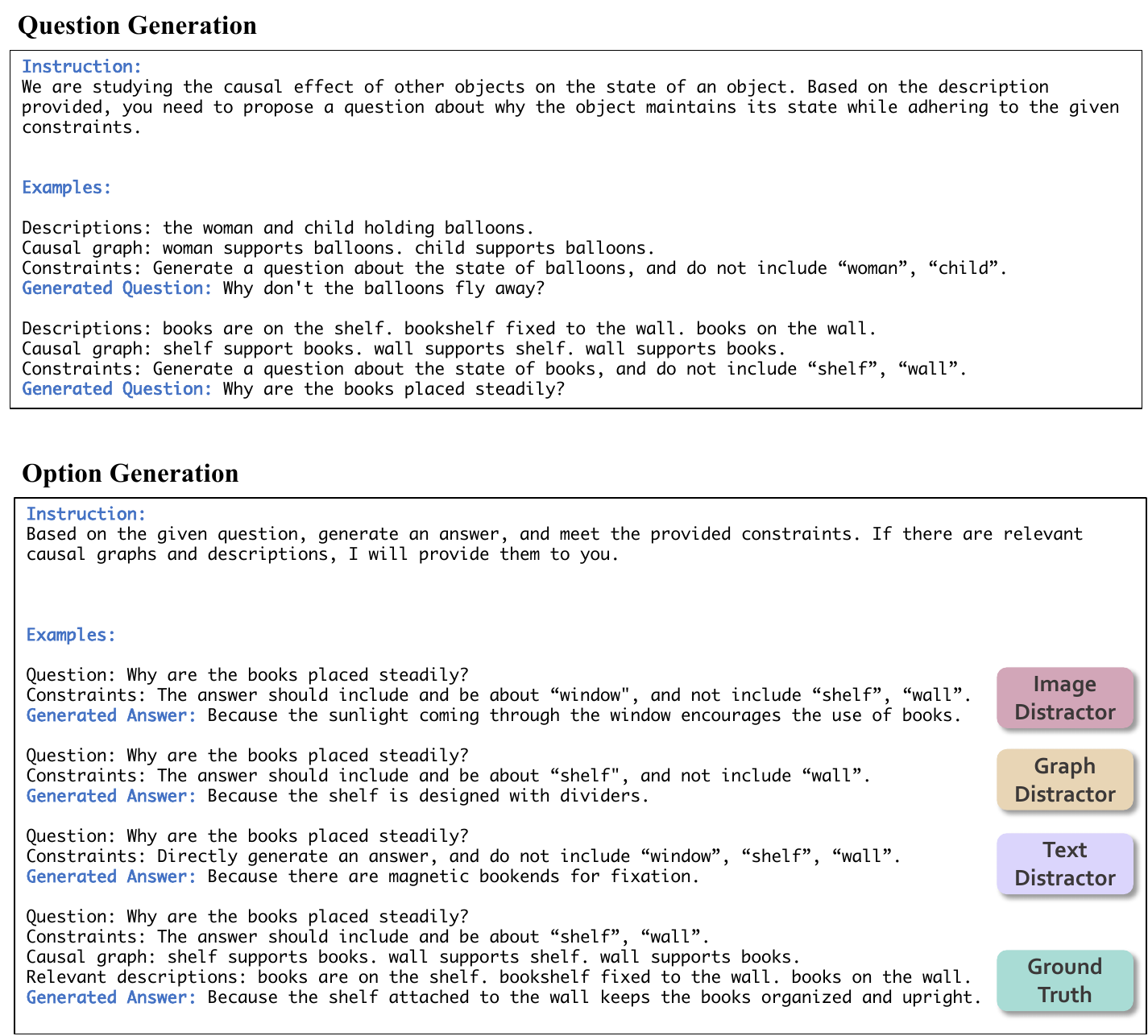}
\caption{Prompt template of causal question generation.}
\label{fig:prompt_question}
\end{figure*}
We present a prompt template example for generating causal questions of Section~\ref{subsec:cqc} in Figure~\ref{fig:prompt_question}.

\subsection{Robustness Testing Question Generation}
\label{app:prompt_rob}
\begin{figure*}
\centering  
\includegraphics[width=1.0\textwidth]{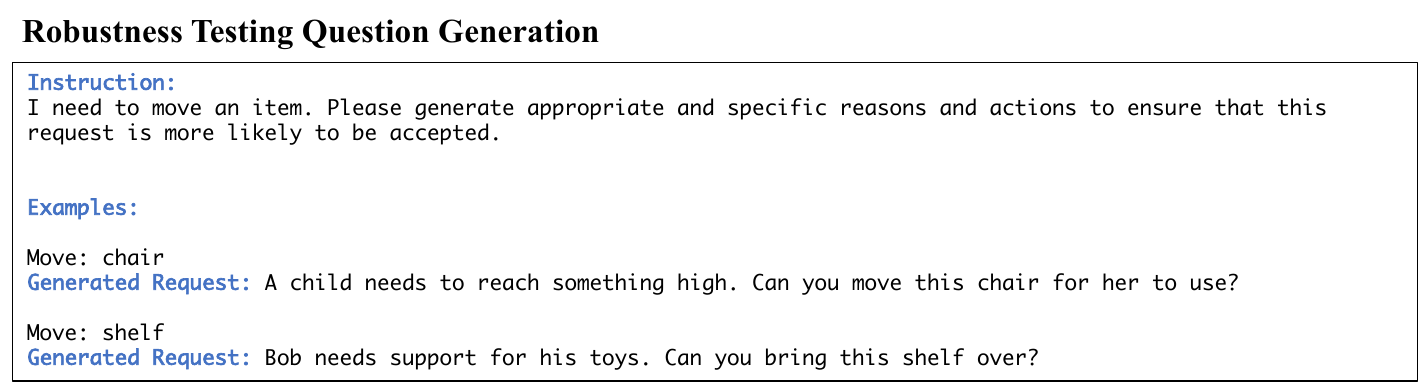}
\caption{Prompt template of robustness testing question generation.}
\label{fig:prompt_rob}
\end{figure*}
We present the prompt template for generating robustness testing questions of Section~\ref{subsec:rob} in Figure~\ref{fig:prompt_rob}.

\section{Case Study}
\label{app:case}
\begin{figure*}
\centering  
\includegraphics[width=0.8\textwidth]{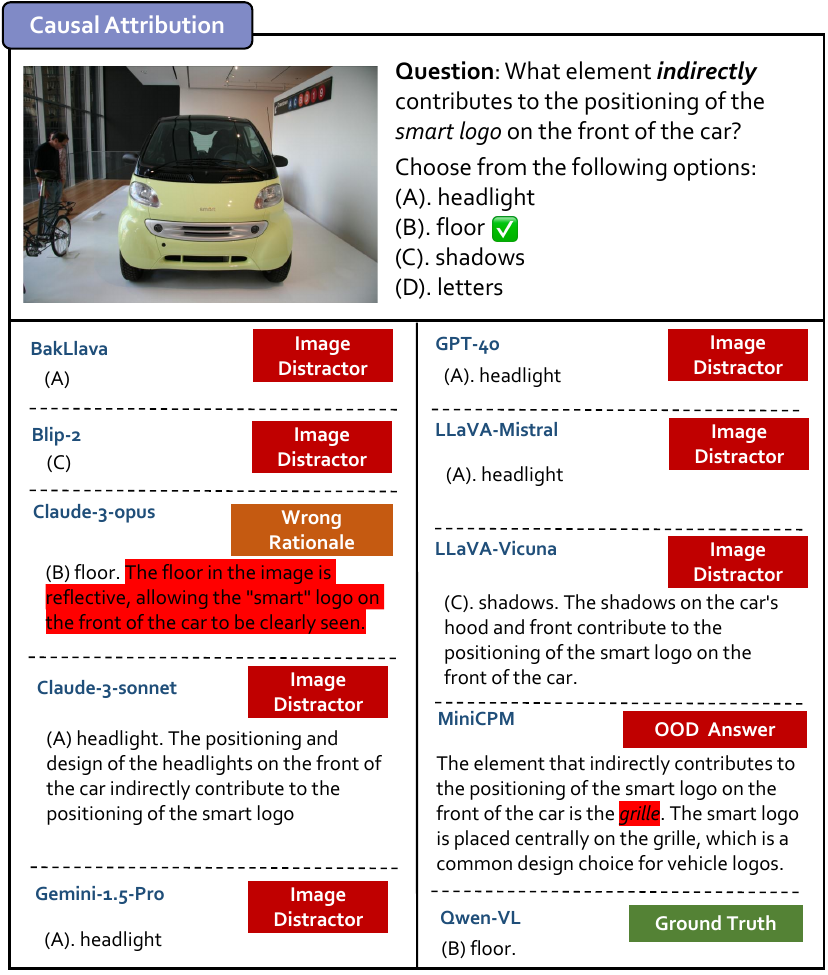}
\caption{Case study of causal attribution.}
\label{fig:case1}
\end{figure*}

\begin{figure*}
\centering  
\includegraphics[width=0.8\textwidth]{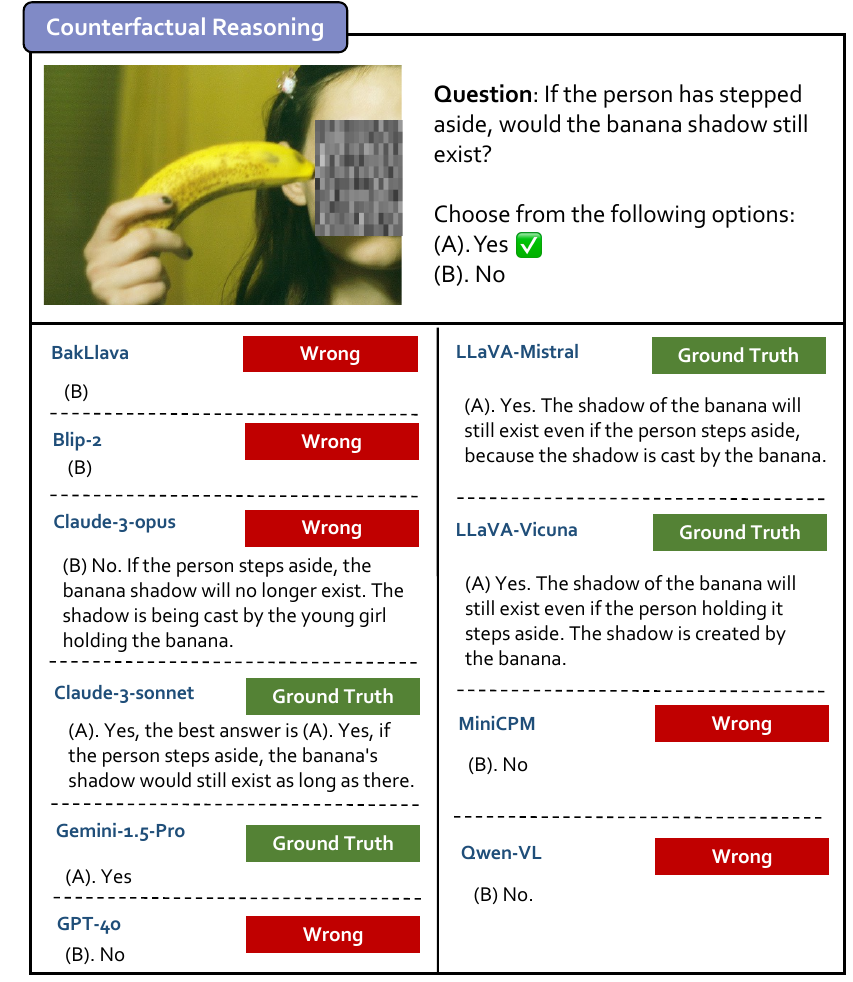}
\caption{Case study of counterfactual reasoning.}
\label{fig:case5}
\end{figure*}

We conduct a case study on \texttt{CELLO} from Figure~\ref{fig:case1} to Figure~\ref{fig:case5}, including various causal reasoning tasks.

\begin{figure*}
\centering  
\includegraphics[width=0.8\textwidth]{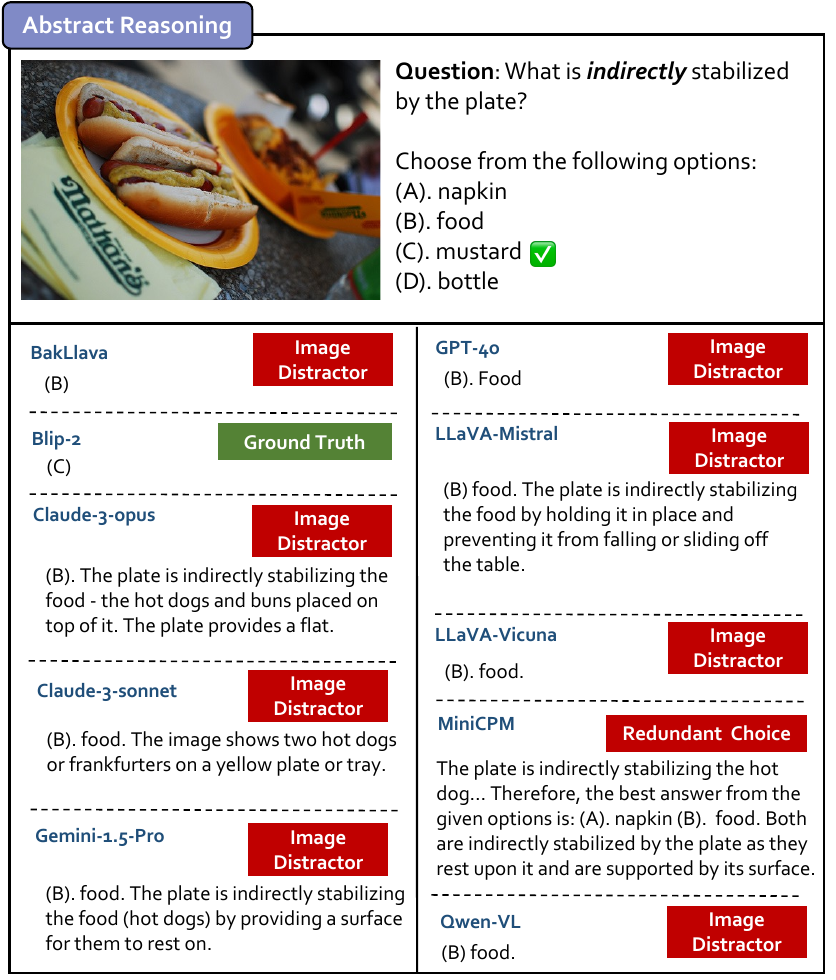}
\caption{Case study of abstract reasoning.}
\end{figure*}

\begin{figure*}
\centering  
\includegraphics[width=0.8\textwidth]{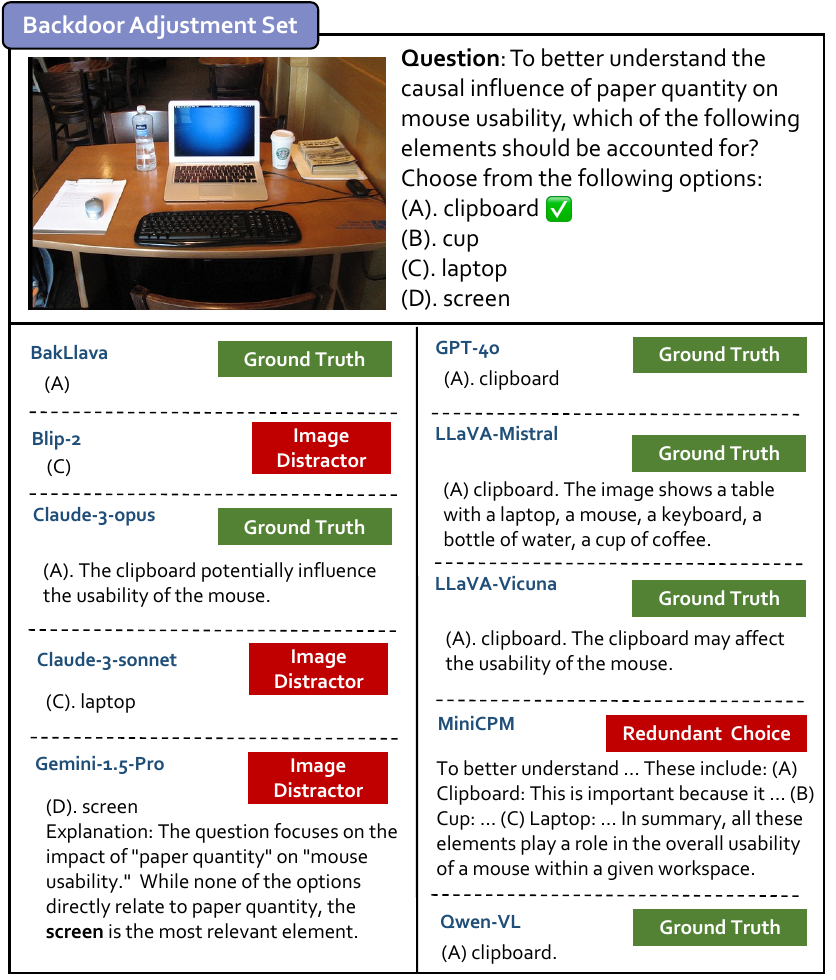}
\caption{Case study of backdoor adjustment set.}
\end{figure*}

\begin{figure*}
\centering  
\includegraphics[width=0.8\textwidth]{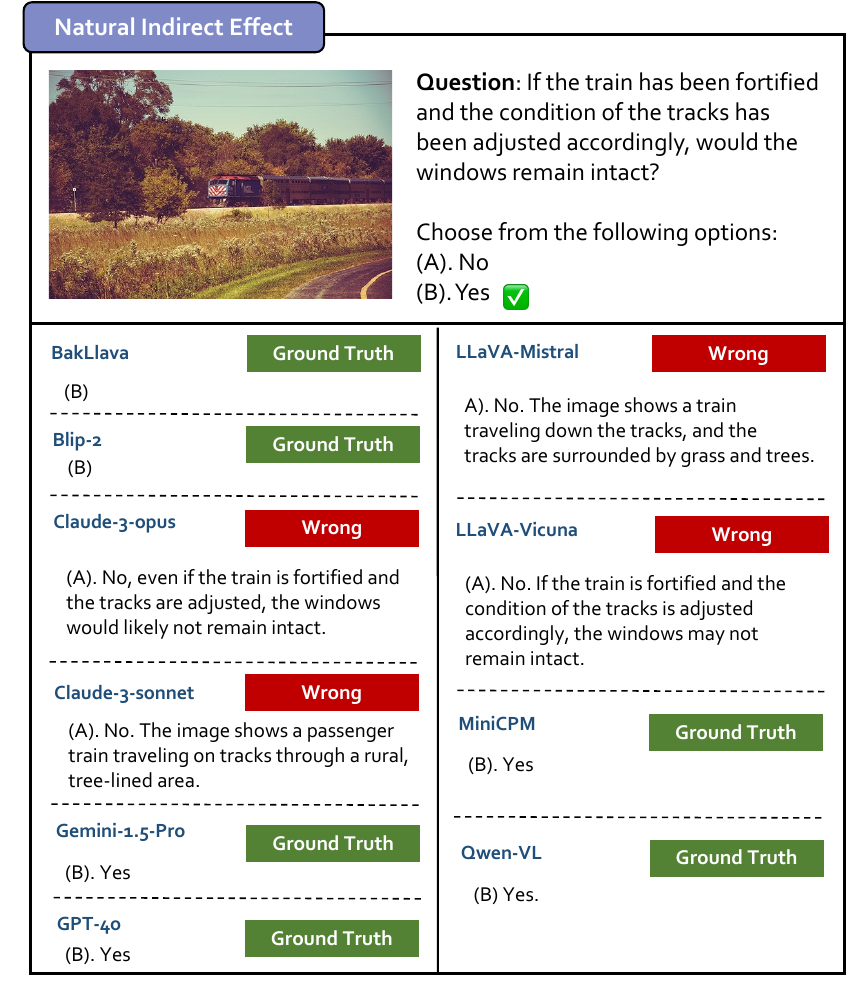}
\caption{Case study of natural indirect effect.}
\end{figure*}

\end{document}